\pdfoutput=1

\documentclass[11pt]{article}

\usepackage[final]{acl}

\usepackage{times}
\usepackage{latexsym}

\usepackage[T1]{fontenc}

\usepackage[utf8]{inputenc}

\usepackage{microtype}

\usepackage{inconsolata}

\usepackage{graphicx}

\usepackage[flushleft]{threeparttable}
\usepackage{colortbl}
\usepackage{tabularx}
\usepackage{arydshln}
\usepackage{booktabs}
\usepackage{diagbox}
\usepackage{subcaption}
\usepackage{multirow} 
\usepackage{pifont}
\usepackage{listings}
\usepackage{hyperref}
\hypersetup{
    colorlinks=true,
    linkcolor=red,
    citecolor=Green,
    filecolor=magenta,      
    urlcolor=magenta,
}

\usepackage{marvosym}
\usepackage{CJKutf8}
\usepackage{makecell}
\usepackage{xspace}
\usepackage{wrapfig}
\usepackage{cuted}
\usepackage{enumitem}

\usepackage[most]{tcolorbox}
\lstdefinestyle{prompt_json}{
  basicstyle=\ttfamily,
  keywordstyle=\color{blue},
  stringstyle=\color{orange},
  commentstyle=\color{green},
  frame=single,
  rulecolor=\color{black},\textit{}
  breakatwhitespace=false,
  breaklines=true,
  captionpos=b,
  keepspaces=true,
  showspaces=false,
  showstringspaces=false,
  showtabs=false,
  tabsize=2
}

\tcbset{
  aibox/.style={
    top=10pt,
    colback=white,
    colframe=NavyBlue,
    colbacktitle=RoyalBlue,
    enhanced,
    center,
    breakable,
    attach boxed title to top left={yshift=-0.1in,xshift=0.15in},
    boxed title style={boxrule=0pt,colframe=white,},
  }
}
\newtcolorbox{AIbox}[2][]{aibox, title=#2,#1}

\usepackage{amsmath}
\usepackage{amssymb}
\usepackage{mathtools}
\usepackage{amsthm}
\usepackage{thmtools}

\usepackage[capitalize,noabbrev]{cleveref}

\usepackage{thm-restate}
\theoremstyle{plain}

\theoremstyle{definition}

\theoremstyle{remark}

\usepackage[textsize=tiny]{todonotes}

\usepackage{xspace}
\newcommand{\gpk}{\text{G-Pass@}$k$\xspace}
\newcommand{\gpkt}{\text{G-Pass@}$k_{\tau}$\xspace}

\title{Are Your LLMs Capable of Stable Reasoning?}

\author{
  Junnan Liu, Hongwei Liu, Linchen Xiao, Ziyi Wang, Kuikun Liu, Songyang Gao \\
  \bf{Wenwei Zhang}, Songyang Zhang\footnotemark[2], Kai Chen\footnotemark[2] \\
  Shanghai Artificial Intelligence Laboratory \\
   {\small \texttt{to.liujn@outlook.com,zhangsongyang@pjlab.org.cn}} \\
  }

\begin{document}
\maketitle
\renewcommand{\thefootnote}{\fnsymbol{footnote}}
\footnotetext[2]{Corresponding authors.}

\begin{abstract}

The rapid advancement of large language models (LLMs) has shown remarkable progress in complex reasoning tasks. 
However, a significant disparity exists between benchmark performances and real-world applications. 
We attribute this gap primarily to current evaluation protocols and metrics, which inadequately capture the full spectrum of LLM capabilities, especially in complex reasoning tasks where both accuracy and consistency are essential. 
In this paper, we introduce \textbf{\gpk}, a novel evaluation metric that continuously assesses model performance across multiple sampling attempts, quantifying both the model's performance potential and its stability. 
Through extensive experiments on various public and newly constructed benchmarks, we employ \gpk in conjunction with state-of-the-art large language models to provide comprehensive insights into their potential capabilities and operational consistency. 
Our findings reveal a significant opportunity to enhance the realistic reasoning abilities of LLMs, underscoring the necessity for more robust evaluation metrics.
Our code and data are available in \url{https://github.com/open-compass/GPassK}.

\end{abstract}

\section{Introduction}

Since the emergence of large language models~(LLMs), complex reasoning, particularly in mathematical problem-solving, has been regarded as the \textit{crown jewel} of LLM capabilities~\citep{numina_math_7b,DBLP:journals/corr/abs-2402-03300}. 
Numerous models have demonstrated remarkable performance on mathematical tasks, from general-purpose models like GPT-4o~\citep{gpt4o}, and the LLaMA series~\citep{llama3_70b_instruct} to specialized models such as DeepSeek-Math~\citep{DBLP:journals/corr/abs-2402-03300} and Qwen-Math~\citep{DBLP:journals/corr/abs-2409-12122}, which excel in complex and high-difficulty mathematical reasoning. 
More recently, long-chain-of-thought (Long-CoT) models like OpenAI-o1~\citep{o1preview}, QwQ~\citep{qwq-32b-preview}, and DeepSeek-R1~\citep{deepseekai2025deepseekr1incentivizingreasoningcapability} have further advanced the state-of-the-art in mathematical problem-solving.

In real-world applications, LLMs typically employ sampling with predefined decoding parameters (\textit{e.g.}, temperature, top-k, top-p, and repetition penalty) to maintain response diversity. 
Users often regenerate responses or initiate new sessions until obtaining satisfactory answers to specific questions. 
However, conventional evaluation metrics for LLMs, such as Greedy Accuracy, Pass@$k$~\citep{DBLP:journals/corr/abs-2107-03374}, Best-of-N (BoN), and Majority Voting, demonstrate significant limitations in measuring real-world performance, particularly regarding long-term consistency. 
While these metrics effectively capture either instantaneous accuracy or peak performance across multiple samples, they inadequately reflect \textbf{output stability}. 
Such instability poses significant challenges for applications requiring reliable and predictable outcomes, highlighting the need for evaluation metrics that effectively balance response diversity with consistent performance and comprehensive analysis of the reasoning stability of current LLMs.

To address these challenges, we introduce \textbf{\gpk}, a novel evaluation metric that simultaneously assesses both reasoning capability and performance consistency. 
The core principle of \gpk lies in evaluating model performance under varying thresholds of correctness, providing a nuanced understanding of model behavior across different stringency levels. 
By integrating measures of both stability and potential, \textbf{\gpk} delivers a comprehensive assessment of an LLM's capabilities, particularly valuable for complex reasoning tasks. 

To thoroughly study the reasoning stability of current LLMs using \textbf{\gpk}, we conduct comprehensive experiments to evaluate various models on multiple mathematical reasoning benchmarks, encompassing different categories of large language models. 
These include general LLMs, mathematics-specialized models, and Chain-of-Thought (CoT) enhanced reasoning models, which are known for their step-by-step reasoning capabilities in complex mathematical problem-solving. 
We also introduce LiveMathBench, a challenging bilingual mathematics benchmark assembled from multiple competitions and examinations to mitigate the impact of data leakage. 

Through systematic comparison between traditional metrics~(Greedy Accuracy, Pass@$k$) and \gpk, our findings reveal distinctive insights into model performance that conventional evaluation approaches may overlook and demonstrate significant limitations in traditional evaluation metrics regarding output stability assessment. 
We observe a substantial performance degradation across all models as the \gpk threshold becomes more stringent, a pattern consistent across both established open-source benchmarks~(MATH~\citep{DBLP:conf/iclr/LightmanKBEBLLS24} and AIME~\citep{AIME2024,AIME2025}) and our proposed LiveMathBench. 
Specifically, the key observations include:
\begin{itemize}[leftmargin=*]
    \item \textbf{Instability in LLM Reasoning}~(\Cref{sec:main_performance}, \Cref{sec:performance_wrt_difficulty}): Both closed-source and open-source models exhibit significant instability when applied to challenging reasoning tasks. Performance drops exceed 50\% in many cases, with the worst instances showing declines of up to 90\%. This issue is particularly significant in high-difficulty questions, underscoring the necessity for further enhancement of current LLMs' ability to address complex questions.
    \item \textbf{Inconsistency Between Expanded Model Scale and Enhanced Reasoning Ability}~(\Cref{sec:main_performance}): Merely scaling up model size does not necessarily enhance stable reasoning capabilities. The expected improvements in performance and stability are not consistently observed.
    \item \textbf{Discrepancy Between Potential and Stability}~(\Cref{sec:main_performance}): There is a noticeable gap between the models' potential capabilities, as measured by their $\text{G-Pass@}k_{\tau\rightarrow 0}$, and their actual stability, reflected in $\text{G-Pass@}k_\tau$. This disparity highlights the need for further research into developing methods that can better harness the theoretical capabilities of these models in practical, stable applications.
\end{itemize}

We also investigate the robustness of the newly proposed metric \gpk, demonstrating that \gpk consistently evaluates model performance across different hyperparameters, guaranteeing its reliability and applicability~(\Cref{sec:performance_wrt_param}, \Cref{sec:performance_wrt_sampling}).
Furthermore, we conduct a preliminary investigation into enhancing the reasoning stability of the model and identifying the reasons behind strong stability.
Our investigation reveals two key findings: 
1) merely relying on SFT does not achieve significant improvement in reasoning stability~(\Cref{sec:performance_wrt_contamination}), and 
2) the incorporation of backtracking and self-reflection mechanisms in o1-like models improves the stability of their reasoning capabilities~(\Cref{sec:long_cot_improve}).

\section{Generalized Metric for LLM Reasoning}

\subsection{Preliminary: \texorpdfstring{Pass@$k$}{Pass@k}}

Pass@$k$ was initially proposed to evaluate the functional correctness of code generated by models~\citep{DBLP:conf/nips/KulalPC0PAL19,DBLP:journals/corr/abs-2107-03374}. 
With the growing application of large language models across diverse reasoning tasks~\citep{DBLP:conf/acl/RajaniMXS19,DBLP:conf/acl/ImaniD023,DBLP:conf/emnlp/GiadikiaroglouL24}, the Pass@$k$ metric has gained increasing recognition~\citep{DBLP:journals/corr/abs-2308-09583,DBLP:conf/iclr/YuJSYLZKLWL24}. 
It provides an effective measure of a model's \textbf{potential performance} in solving complex questions. Pass@$k$ represents the probability of generating \textbf{at least one correct solution} within $k$ attempts for each question, as defined by the formula:
\begin{equation} \label{eq:pass_at_k}
    \text{Pass@}k = \mathbb{E}_{\text{Questions}} \left[ 1 - \frac{{{n - c} \choose k}}{{n \choose k}} \right],
\end{equation}
where $n$ denotes the total number of generations in the reasoning task, and $c$ is the number of correct solutions among them. 
Intuitively, \Cref{eq:pass_at_k} calculates the expected proportion of questions for which at least one correct solution is obtained. 
In practice, $n$ is typically set equal to $k$, primarily to minimize inference costs.

\subsection{Generalized Metric: \texorpdfstring{\gpk}{G-Pass@k}}

While Pass@$k$ provides an indication of a model's performance potential, it does not consider the stability of the model's reasoning across diverse attempts. 
To evaluate both the potential and stability of a model, we propose a generalized metric called \textbf{\gpk}. 
Simply put, \gpk assesses the stability of a reasoning model by measuring the probability of consistently generating correct solutions across all attempts.

\paragraph{Definition of \texorpdfstring{\gpk}{G-Pass@k} and \texorpdfstring{\gpkt}{G-Pass@k_\tau}. }
Let $p^*$ denote the latent success probability of a model generating correct solutions. 
Assuming each generation is independent and identically distributed (i.i.d.), the probability of obtaining $m$ correct solutions follows a \textit{binomial distribution}:
\begin{equation}
    m \sim B(n, p^*).
\end{equation}
Since $p^*$ is typically inaccessible, we use a \textit{hypergeometric distribution} to approximate the \textit{binomial distribution}:
\begin{equation}
    \lim_{n \rightarrow \infty} H(m;k, c, n) \rightarrow B(m;n,p^*). 
\end{equation}
Therefore, \gpk can be defined as~\citep{DBLP:journals/corr/abs-2406-12045}:
\begin{equation} \label{eq:strict_k_at_pass}
    \text{G-Pass@}k = \mathbb{E}_{\text{Questions}} \left[ \frac{{c \choose k}}{{n \choose k}} \right].
\end{equation}
Here, $n$ denotes the total number of generations per question, and $c$ signifies the number of generations that yield correct solutions.
Considering the stringent requirement in \Cref{eq:strict_k_at_pass}, we draw inspiration from the mean Average Precision (mAP) metric~\citep{DBLP:journals/ijcv/EveringhamGWWZ10} in the object detection task to introduce a tolerance threshold $\tau \in (0.0, 1.0]$, leading to the definition of \gpkt:
\begin{equation} \label{eq:k_at_pass}
    \text{G-Pass@}k_{\tau} = \mathbb{E}_{\text{Questions}} \left[ \sum_{j= \lceil \tau \cdot k \rceil}^{c} \frac{{c \choose j} \cdot {{n - c} \choose {k - j}}}{{n \choose k}} \right],
\end{equation}
where $\lceil \tau \cdot k \rceil$ denotes the smallest integer greater than or equal to $\tau \cdot k$. 
Conceptually, for $\tau < 1.0$, there is flexibility to allow up to $k - \lceil \tau \cdot k \rceil$ incorrect solutions within the $k$ generations. 
In conclusion, we leverage the hypergeometric distribution for sampling without replacement to approximate the binomial distribution for sampling with replacement. 
Such approximations tend to have smaller errors when $n$ is sufficiently large ($k \ll n$). 
Further details are provided in \Cref{app:k_at_pass_estimation}.

\paragraph{Pass@$k$ as a Special Case of \gpkt. }
Pass@$k$ can be regarded as a special case of \gpkt, as formalized in the following theorem:
\begin{restatable}{theorem}{them} \label{theo:k_at_pass_equal_pass_at_k}
    Pass@$k$ is a special case of \gpk as $\tau$ approaches 0, formally expressed as:
    \begin{equation}
        \lim_{\tau \rightarrow 0} \sum_{j= \lceil \tau \cdot k \rceil}^{c} \frac{{c \choose j} \cdot {{n - c} \choose {k - j}}}{{n \choose k}} = 1 - \frac{{{n - c} \choose k}}{{n \choose k}}.
    \end{equation}
\end{restatable}
The proof is provided in \Cref{app:prof_k_at_pass_equal_pass_at_k}. 

\paragraph{Definition of m\gpk. }
Recall that when the threshold $\tau$ is low, G-Pass@$k_\tau$ primarily reflects the model's performance potential. 
Conversely, at higher $\tau$ values, G-Pass@$k_\tau$ evaluates the model's stability, \textit{i.e.}, its level of mastery over the question. 
Thus, G-Pass@$k_\tau$ enables the continuous assessment of both performance potential and stability.
We further define \textbf{mG-Pass@}$k$ as follows:
\begin{equation}
    \begin{aligned}
        \text{mG-Pass@}k_{\tau} &= 2\int_{0.5}^{1.0} \text{G-Pass@}k_{\tau} d \tau \\
        & = \frac{2}{k} \sum_{i= \lceil 0.5 \cdot k \rceil + 1}^{k} \text{G-Pass@}k_{\frac{i}{k}}.
    \end{aligned}
\end{equation}
Intuitively, mG-Pass@$k$ provides an interpolated estimate of the area under the curve of \gpkt for $\tau \in [0.5, 1.0]$, serving as a comprehensive metric that integrates all \gpkt values within a certain range. 
For models that are both optimal and stable, the mG-Pass@$k$ value should approach $1$.

\begin{table*}[t]
    \centering
    \caption{Performance of models on LiveMathBench. We perform $48$ runs and report results of greedy accuracy, and G-Pass@$16_{\{0.5, 0.75, 1.0\}}$ and mG-Pass@$16$. A more detailed performance can be found in \Cref{tab:full_main_performance} at \Cref{app:full_performance_livemath}.}
    \label{tab:main_performance}
    \resizebox{.9\textwidth}{!}{
        \begin{threeparttable}[b]
        \begin{tabular}{lccccc}
            \toprule
            \multirow{2}{*}{\textbf{LLMs}} & \multirow{2}{*}{\textbf{Greedy}} & \multicolumn{4}{c}{\textbf{G-Pass@$16$}~(\Cref{eq:k_at_pass}) / \%} \\
            \cmidrule(lr){3-6}
            & & \textbf{G-Pass@$16_{0.5}$} & \textbf{G-Pass@$16_{0.75}$} & \textbf{G-Pass@$16_{1.0}$} &  \textbf{mG-Pass@$16$} \\
            \midrule
            \multicolumn{6}{c}{\textbf{General LLMs}} \\
            \midrule
            {\small Llama-3.1-8B-Instruct} & \cellcolor{Purple!24.0} $24.0$ & $18.2$ & $11.3$ & $4.5$ & \cellcolor{Cerulean!10.4} $10.4$ \\ 
            {\small Yi-1.5-34B-Chat} & \cellcolor{Purple!24.8} $24.8$ & $18.6$ & $11.3$ & $6.0$ & \cellcolor{Cerulean!11.0} $11.0$ \\
            {\small Gemma-2-27b-it} & \cellcolor{Purple!26.9} $26.9$ & $23.5$ & $17.8$ & $12.7$ & \cellcolor{Cerulean!17.3} $17.3$ \\
            {\small Llama-3.1-70B-Instruct} & \cellcolor{Purple!29.8} $29.8$ & $30.0$ & $22.2$ & $12.5$ & \cellcolor{Cerulean!20.8} $20.8$ \\
            {\small Qwen2.5-7B-Instruct} & \cellcolor{Purple!37.0} $37.0$ & $36.5$ & $27.2$ & $16.0$ & \cellcolor{Cerulean!25.8} $25.8$ \\
            {\small DeepSeek-V2.5-1210} & \cellcolor{Purple!38.7} $38.7$ & $38.9$ & $27.9$ & $17.3$ & \cellcolor{Cerulean!26.7} $26.7$ \\
            {\small Llama-3.3-70B-Instruct} & \cellcolor{Purple!40.3} $40.3$ & $36.2$ & $28.9$ & $19.1$ & \cellcolor{Cerulean!27.5} $27.5$ \\
            {\small GPT-4o-2024-11-20} \tnote{\dag} & \cellcolor{Purple!44.8} $44.8$ & $41.9$ & $32.9$ & $22.2$ & \cellcolor{Cerulean!31.6} $31.6$ \\
            {\small InternLM3-8B-Instruct} & \cellcolor{Purple!44.5} $44.5$ & $43.0$ & $35.4$ & $23.0$ & \cellcolor{Cerulean!33.6} $33.6$ \\ 
            {\small Claude-3.5-Sonnet} \tnote{\dag} & \cellcolor{Purple!46.7} $46.7$ & $44.1$ & $36.2$ & $26.6$ & \cellcolor{Cerulean!35.3} $35.3$ \\
            {\small Mistral-Large-Instruct-2411} & \cellcolor{Purple!41.6} $41.6$ & $39.4$ & $37.1$ & $32.9$ & \cellcolor{Cerulean!36.4} $36.4$ \\
            {\small Qwen2.5-72B-Instruct} & \cellcolor{Purple!51.7} $51.7$ & $47.3$ & $39.6$ & $29.0$ & \cellcolor{Cerulean!37.8} $37.8$\\
            {\small Qwen2.5-32B-Instruct} & \cellcolor{Purple!50.8} $50.8$ & $48.3$ & $39.5$ & $28.6$ & \cellcolor{Cerulean!38.1} $38.1$ \\
            {\small Gemini-1.5-Pro-Latest} \tnote{\dag} & \cellcolor{Purple!59.1} $59.1$ & $55.9$ & $47.3$ & $31.0$ & \cellcolor{Cerulean!44.3} $44.3$ \\
            {\small DeepSeek-V3.0-Chat} \tnote{\dag} & \cellcolor{Purple!55.0} $55.0$ & $59.5$ & $49.9$ & $35.0$ & \cellcolor{Cerulean!47.9} $47.9$ \\
            \midrule
            \multicolumn{6}{c}{\textbf{Mathematical Reasoning LLMs}} \\
            \midrule
            {\small DeepSeek-Math-7B-RL} & \cellcolor{Purple!23.5} $23.5$ & $19.8$ & $14.0$ & $9.7$ & \cellcolor{Cerulean!13.7} $13.7$ \\
            {\small NuminaMath-72B-CoT} & \cellcolor{Purple!40.8} $40.8$ & $34.0$ & $27.1$ & $14.2$ & \cellcolor{Cerulean!25.0} $25.0$ \\
            {\small Qwen2.5-Math-7B-Instruct} & \cellcolor{Purple!44.1} $44.1$ & $44.1$ & $38.3$ & $28.1$ & \cellcolor{Cerulean!36.6} $36.6$ \\
            {\small Qwen2.5-Math-72B-Instruct} & \cellcolor{Purple!57.6} $57.6$ & $52.7$ & $45.4$ & $27.9$ & \cellcolor{Cerulean!42.3} $42.3$ \\
            \midrule
            \multicolumn{6}{c}{\textbf{O1-like Reasoning LLMs}} \\
            \midrule
            {\small Skywork-o1} & \cellcolor{Purple!45.4} $45.4$ & $39.3$ & $31.9$ & $21.7$ & \cellcolor{Cerulean!30.4} $30.4$ \\ 
            {\small QwQ-32B-Preview} & \cellcolor{Purple!72.7} $72.7$ & $74.9$ & $65.8$ & $40.1$ & \cellcolor{Cerulean!61.2} $61.2$ \\
            {\small OpenAI o1-mini} \tnote{\dag} & \cellcolor{Purple!74.1} $74.1$ \tnote{\ddag} & $76.3$ & $67.3$ & $48.3$ & \cellcolor{Cerulean!64.8} $64.8$ \\
            {\small DeepSeek Distill Qwen-32B} & \cellcolor{Purple!67.7} $67.7$ & $81.2$ & $72.3$ & $54.5$ & \cellcolor{Cerulean!69.7} $69.7$ \\
            {\small DeepSeek Distill LLama-70B} & \cellcolor{Purple!74.8} $74.8$ & $80.8$ & $73.0$ & $53.0$ & \cellcolor{Cerulean!69.7} $69.7$ \\
            {\small DeepSeek R1} \tnote{\dag} & \cellcolor{Purple!81.1} $81.1$ &  $83.6$ & $79.1$ & $69.5$ & \cellcolor{Cerulean!77.6} $77.6$ \\
            \bottomrule
        \end{tabular}
        \begin{tablenotes}
            \item[\dag] API-based LLMs.
            \item[\ddag] OpenAI o1 series model does not provide an optional temperature parameter, so we chose the average accuracy of 20 generations as the proxy for greedy accuracy.
        \end{tablenotes}
        \end{threeparttable}
    }
\end{table*}

\section{Stability of LLM Reasoning}

In this section, we perform extensive experiments on the mathematical reasoning task to evaluate and analyze the reasoning stability of current LLMs, given the prevalence and ease of verification of the mathematical reasoning task.

\begin{table*}[t]
    \centering
    \caption{Performance of models on MATH500 and AIME2024. Aligning with experiments on LiveMathBench, we also perform $48$ runs and report results of greedy accuracy, G-Pass@$16_{\{0.5, 0.75, 1.0\}}$, and mG-Pass@$16$. More detailed results are available in \Cref{tab:full_public_performance} at \Cref{app:full_performance_public}.}
    \label{tab:public_performance}
    \resizebox{.9\textwidth}{!}{
        \begin{threeparttable}[b]
        \begin{tabular}{lccccc}
            \toprule
            \multirow{2}{*}{\textbf{LLMs}} & \multirow{2}{*}{\textbf{Greedy}} & \multicolumn{4}{c}{\textbf{G-Pass@$16$}~(\Cref{eq:k_at_pass}) / \%} \\
            \cmidrule(lr){3-6}
            & & \textbf{G-Pass@$16_{0.5}$} & \textbf{G-Pass@$16_{0.75}$} & \textbf{G-Pass@$16_{1.0}$} & \textbf{mG-Pass@$16$} \\
            \midrule
            \multicolumn{6}{c}{\textbf{MATH500-L5}} \\
            \midrule
            {\small Qwen2.5-7B-Instruct} & \cellcolor{Purple!56.7}  $56.0$ & $54.8$ & $43.3$ & $28.0$ & \cellcolor{Cerulean!41.5} $41.5$ \\
            {\small Llama-3.3-70B-Instruct} & \cellcolor{Purple!54.5}  $54.5$ & $55.4$ & $49.5$ & $35.0$ & \cellcolor{Cerulean!44.9} $47.3$ \\
            {\small Mistral-Large-Instruct-2411} & \cellcolor{Purple!55.2}  $55.2$ & $52.3$ & $51.2$ & $45.6$ & \cellcolor{Cerulean!50.1} $50.1$ \\
            {\small Qwen2.5-72B-Instruct} & \cellcolor{Purple!63.4}  $63.4$ & $62.5$ & $54.4$ & $44.9$ & \cellcolor{Cerulean!53.1} $53.1$\\
            {\small Qwen2.5-32B-Instruct} & \cellcolor{Purple!64.2}  $64.2$ & $66.6$ & $59.4$ & $41.0$ & \cellcolor{Cerulean!55.6} $55.6$ \\
            \hdashline
            {\small Qwen2.5-Math-7B-Instruct} & \cellcolor{Purple!65.7}  $65.7$ & $65.0$ & $62.2$ & $57.6$ & \cellcolor{Cerulean!61.5} $61.5$ \\
            {\small Qwen2.5-Math-72B-Instruct} & \cellcolor{Purple!71.6}  $71.6$ & $64.9$ & $59.4$ & $46.0$ & \cellcolor{Cerulean!57.6} $57.6$ \\
            \hdashline
            {\small Skywork-o1} & \cellcolor{Purple!61.2}  $61.2$ & $56.5$ & $52.2$ & $42.9$ & \cellcolor{Cerulean!50.7} $50.7$ \\
            {\small QwQ-32B-Preview} & \cellcolor{Purple!83.6}  $83.6$ & $87.2$ & $78.8$ & $57.4$ & \cellcolor{Cerulean!75.6} $75.6$ \\
            {\small DeepSeek Distill Qwen-32B} & \cellcolor{Purple!83.6} $83.6$ & $89.9$ & $83.8$ & $70.4$ & \cellcolor{Cerulean!81.9} $81.9$ \\
            \midrule
            \multicolumn{6}{c}{\textbf{AIME2024-45}} \\
            \midrule
            {\small Qwen2.5-32B-Instruct} & \cellcolor{Purple!11.1}  $11.1$ & $7.1$ & $3.4$ & $2.2$ & \cellcolor{Cerulean!4.8} $3.7$ \\
            {\small Qwen2.5-7B-Instruct} & \cellcolor{Purple!11.1}  $11.1$ & $8.9$ & $8.1$ & $4.7$ & \cellcolor{Cerulean!7.5} $7.5$ \\
            {\small Mistral-Large-Instruct-2411} & \cellcolor{Purple!13.3}  $13.3$ & $10.4$ & $6.8$ & $2.4$ & \cellcolor{Cerulean!6.1} $6.1$ \\
            {\small Qwen2.5-72B-Instruct} & \cellcolor{Purple!13.3}  $13.3$ & $13.7$ & $12.9$ & $7.5$ & \cellcolor{Cerulean!11.7} $11.7$ \\
            {\small Llama-3.3-70B-Instruct} & \cellcolor{Purple!22.2}  $22.2$ & $25.3$ & $18.2$ & $6.9$ & \cellcolor{Cerulean!16.4} $16.4$ \\
            \hdashline
            {\small Qwen2.5-Math-7B-Instruct} & \cellcolor{Purple!11.1}  $11.1$ & $4.6$ & $2.6$ & $2.2$ & \cellcolor{Cerulean!2.8} $2.8$ \\
            {\small Qwen2.5-Math-72B-Instruct} & \cellcolor{Purple!20.0}  $20.0$ & $18.7$ & $16.2$ & $6.7$ & \cellcolor{Cerulean!14.1} $14.1$ \\
            \hdashline
            {\small Skywork-o1} & \cellcolor{Purple!11.1}  $11.1$ & $11.2$ & $10.3$ & $1.5$ & \cellcolor{Cerulean!8.2} $8.2$ \\
            {\small QwQ-32B-Preview} & \cellcolor{Purple!44.4}  $44.4$ & $41.0$ & $28.6$ & $8.1$ & \cellcolor{Cerulean!24.7} $24.7$ \\
            {\small OpenAI o1-mini} \tnote{\dag} & \cellcolor{Purple!60.3}  $60.3$ \tnote{\ddag} & $62.2$ & $53.3$ & $15.6$ & \cellcolor{Cerulean!43.1} $43.1$ \\
            {\small DeepSeek Distill Qwen-32B} & \cellcolor{Purple!62.2} $62.2$ & $77.0$ & $66.5$ & $31.3$ & \cellcolor{Cerulean!59.3} $59.3$ \\
            \bottomrule
        \end{tabular} 
        \begin{tablenotes}
            \item[\dag] API-based LLMs.
            \item[\ddag] OpenAI o1 series model does not provide an optional temperature parameter, so we chose the average accuracy of 20 generations as greedy accuracy.
        \end{tablenotes}
        \end{threeparttable}
    }
\end{table*}

\subsection{Setup}

\subsubsection{LLMs}

We evaluate various LLMs recognized for their strong mathematical reasoning capabilities, including three categories: general LLMs, mathematical LLMs, and o1-like LLMs.
For the details, please refer to \Cref{app:evaluated_llm}.

\subsubsection{Data}

In all experiments, we involve several benchmarks, which include the newly constructed benchmark named \textbf{LiveMathBench} and publicly available benchmarks, MATH500-L5, AIME2024-45, and AIME2025. 
For more detailed information about benchmarks, please refer to \Cref{app:data_details}.

\subsection{Main Performance} \label{sec:main_performance}

\Cref{tab:main_performance} demonstrates the performance on LiveMathBench and \Cref{tab:public_performance} demonstrates the performance on MATH500-L5 and AIME2024-45. 
More experimental results can be found in \Cref{app:additional_experimental_results}. 
From the results, we derive the following observations:

\paragraph{1) Reasoning Ability Still Needs to be Properly Evaluated.}
While most models demonstrate relatively strong performance in terms of Greedy Accuracy and Pass@$16$, their effectiveness significantly declines when assessed using the \gpk metric. 
Specifically, when $\tau$ is set to 1.0, indicating that the model must generate accurate solutions in all 16 attempts, almost all models experience a drastic decline in performance. 
Notably, even the most robust model in \Cref{tab:main_performance}, DeepSeek R1, displays a $14.3\%$ decline, diminishing from $81.1\%$ to $69.5\%$. 
Even when $\tau$ is relaxed to 0.5, which requires only half of the generated solutions to be correct, General LLMs, Mathematical Reasoning LLMs, and o1-like Reasoning LLMs still experience a decrease in performance. 
This indicates that, under demanding conditions, most models struggle to maintain consistency in their reasoning abilities across multiple samples, irrespective of whether the criteria are strict or lenient. 
These findings underscore the necessity for more rigorous evaluations of models' reasoning capabilities, particularly in scenarios that require consistent and reliable performance across multiple instances. 
The current evaluation metrics, often reliant on single-shot greedy decoding, may not fully capture the real robustness and stability of these models in real-world applications.

\begin{table*}[t]
    \centering
    \caption{Performance on CCEE and WLPM. The table shows the decreasing trend of \textbf{Greedy} \textit{w.r.t.} \textbf{Pass@$16$} and \textbf{G-Pass@$16_{1.0}$} \textit{w.r.t.} \textbf{Greedy}, which are marked with colors of different transparency.}
    \label{tab:difficulty_performance}
    \resizebox{\textwidth}{!}{
        \begin{tabular}{lcccccc}
            \toprule
            \multirow{2}{*}{\textbf{LLMs}} & \multicolumn{3}{c}{\textbf{CCEE}} & \multicolumn{3}{c}{\textbf{WLPMC}} \\
            \cmidrule(lr){2-4} \cmidrule(lr){5-7}
            & \textbf{G-Pass@$16_{\rightarrow 0}$} & $\searrow$ \textbf{Greedy} & $\searrow$ \textbf{G-Pass@$16_{1.0}$} 
            & \textbf{G-Pass@$16_{\rightarrow 0}$} & $\searrow$ \textbf{Greedy} & $\searrow$ \textbf{G-Pass@$16_{1.0}$} \\
            \midrule
            Llama-3.3-70B-Instruct 
            & $75.9$ 
            & \cellcolor{orange!22.2}$59.0_{\downarrow22.2}$ 
            & \cellcolor{orange!40.3}$35.2_{\downarrow40.3}$ 
            & $42.1$ 
            & \cellcolor{orange!78.3}$9.1_{\downarrow78.3}$ 
            & \cellcolor{orange!100.0}$\sim0.0_{\downarrow100.0}$ \\ 
            
            Mistral-Large-Instruct-2411 
            & $71.5$ 
            & \cellcolor{orange!11.0}$63.6_{\downarrow11.0}$ 
            & \cellcolor{orange!17.7}$52.3_{\downarrow17.7}$ 
            & $18.2$ 
            & \cellcolor{orange!49.8}$9.1_{\downarrow49.8}$ 
            & \cellcolor{orange!32.9}$6.1_{\downarrow32.9}$ \\ 
            
            DeepSeek-V3.0-Chat 
            & $84.6$ 
            & \cellcolor{orange!19.3}$68.2_{\downarrow19.3}$ 
            & \cellcolor{orange!21.8}$53.3_{\downarrow21.8}$ 
            & $56.8$ 
            & \cellcolor{orange!68.0}$18.2_{\downarrow68.0}$ 
            & \cellcolor{orange!63.7}$6.6_{\downarrow63.7}$ \\ 
            
            Qwen2.5-72B-Instruct 
            & $80.3$ 
            & \cellcolor{orange!9.4}$72.7_{\downarrow9.4}$ 
            & \cellcolor{orange!21.7}$56.9_{\downarrow21.7}$ 
            & $50.4$ 
            & \cellcolor{orange!63.9}$18.2_{\downarrow63.9}$ 
            & \cellcolor{orange!77.5}$4.1_{\downarrow77.5}$ \\ 
            
            Gemini-1.5-Pro-Latest 
            & $81.8$ 
            & \cellcolor{orange!16.6}$68.2_{\downarrow16.6}$ 
            & \cellcolor{orange!22.1}$53.1_{\downarrow22.1}$ 
            & $60.0$ 
            & \cellcolor{orange!40.0}$36.4_{\downarrow40.0}$ 
            & \cellcolor{orange!88.2}$4.3_{\downarrow88.2}$ \\ 
            
            GPT-4o 
            & $79.4$ 
            & \cellcolor{orange!22.3}$61.7_{\downarrow22.3}$ 
            & \cellcolor{orange!32.3}$41.8_{\downarrow32.3}$ 
            & $29.9$ 
            & \cellcolor{orange!39.1}$18.2_{\downarrow39.1}$ 
            & \cellcolor{orange!78.0}$4.0_{\downarrow78.0}$ \\ 
            \hdashline
            
            
            Qwen2.5-Math-7B-Instruct 
            & $80.5$ 
            & \cellcolor{orange!21.0}$63.6_{\downarrow21.0}$ 
            & \cellcolor{orange!22.5}$49.3_{\downarrow22.5}$ 
            & $43.8$ 
            & \cellcolor{orange!58.5}$18.2_{\downarrow58.5}$ 
            & \cellcolor{orange!96.2}$0.7_{\downarrow96.2}$ \\
            
            Qwen2.5-Math-72B-Instruct 
            & $84.1$ 
            & \cellcolor{orange!8.0}$77.3_{\downarrow8.0}$ 
            & \cellcolor{orange!30.4}$53.8_{\downarrow30.4}$ 
            & $47.0$ 
            & \cellcolor{orange!41.9}$27.3_{\downarrow41.9}$ 
            & \cellcolor{orange!33.3}$18.2_{\downarrow33.3}$ \\ 
            
            \hdashline
            QwQ-32B-Preview 
            & $92.3$ 
            & \cellcolor{orange!6.3}$86.4_{\downarrow6.3}$ 
            & \cellcolor{orange!36.3}$55.0_{\downarrow36.3}$ 
            & $88.4$ 
            & \cellcolor{orange!69.1}$27.3_{\downarrow69.1}$ 
            & \cellcolor{orange!66.9}$11.7_{\downarrow66.9}$ \\
            
            \bottomrule
        \end{tabular}
    }
\end{table*}

\paragraph{2) Increasing Model Size May Not Significantly Enhance Robustness. }
A comparison of models within the same series, such as Qwen2.5-32B-Instruct and Qwen2.5-72B-Instruct, reveals that despite a more than twofold difference in model size, their performance is similar across various metrics and datasets. 
For example, on both LiveMathBench and existing open-source datasets, the difference in \textit{Greedy Accuracy} and \textit{mG-Pass}@k between these two models is within two percentage points. 
Additionally, in the larger LLMs like Mistral-Large-Instruct-2411~(123B), although the scale has increased further, performance and stability have actually declined compared to Qwen2.5-72B-Instruct. 
This suggests that for certain tasks, particularly those requiring deep understanding and logical reasoning, mere parameter expansion may not yield the expected gains in performance or stability. 
Conversely, the current base model still holds untapped potential, and improved training paradigms or test-time scaling methods should be proposed to enhance its reasoning performance and ability, rather than merely increasing the model size~\citep{DBLP:journals/corr/abs-2408-03314}.

\begin{figure}[t]
    \centering
    \includegraphics[width=\linewidth]{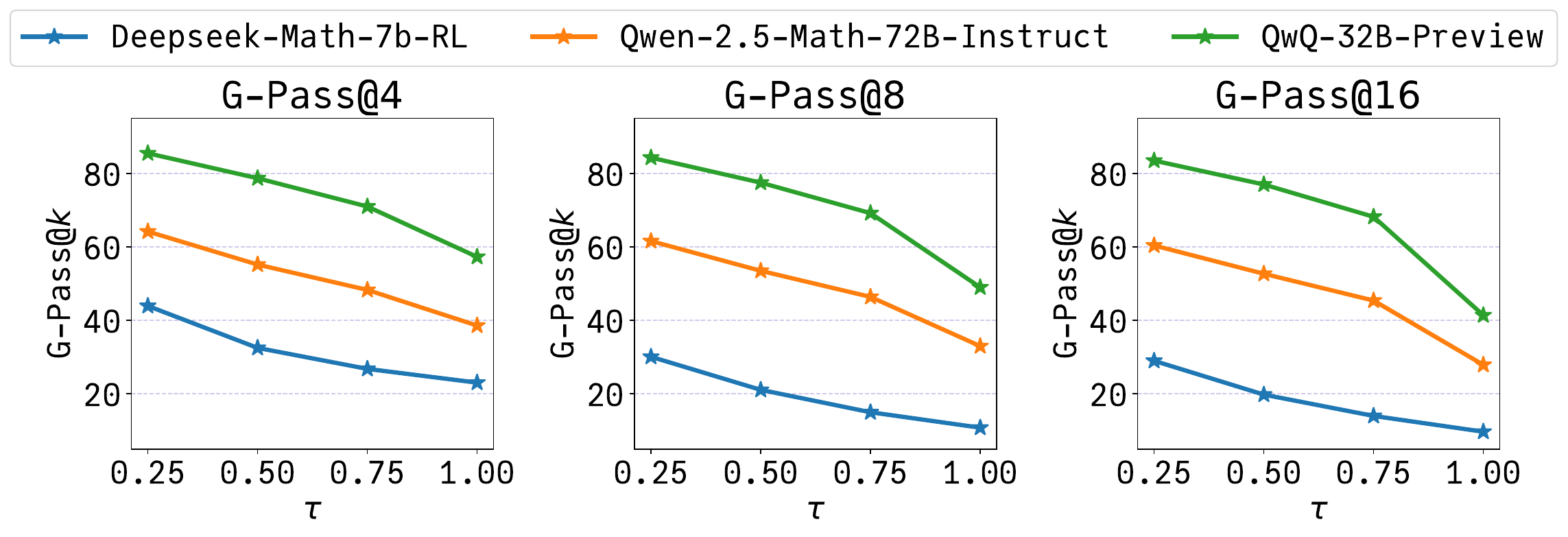}
    \caption{Illustration of \gpk \textit{w.r.t.} different values of $k$, where $k=\{4,8,16\}$ on MATH500-L5.}
    \label{fig:g-pass-at-k-wrt-k}
\end{figure}

\paragraph{3) Significant Gap Between Theoretical Performance Potential and Actual Stability.}
In evaluating model performance, we observe a notable gap between the theoretical upper limit (\textit{G-Pass}@16$_{\tau \rightarrow 0}$), the actual performance (\textit{Greedy Accuracy}), and the stability across multiple samples (\textit{G-Pass}@16$_{\tau=1.0}$). 
As evident from the main performance, while models theoretically possess high potential performance, their actual performance in practical applications falls short of this optimal level, particularly in terms of output stability. 

\subsection{Performance \textit{w.r.t.} \texorpdfstring{\gpkt}{G-Pass@k} Settings} \label{sec:performance_wrt_param}

\begin{figure}[t]
    \centering
    \includegraphics[width=\linewidth]{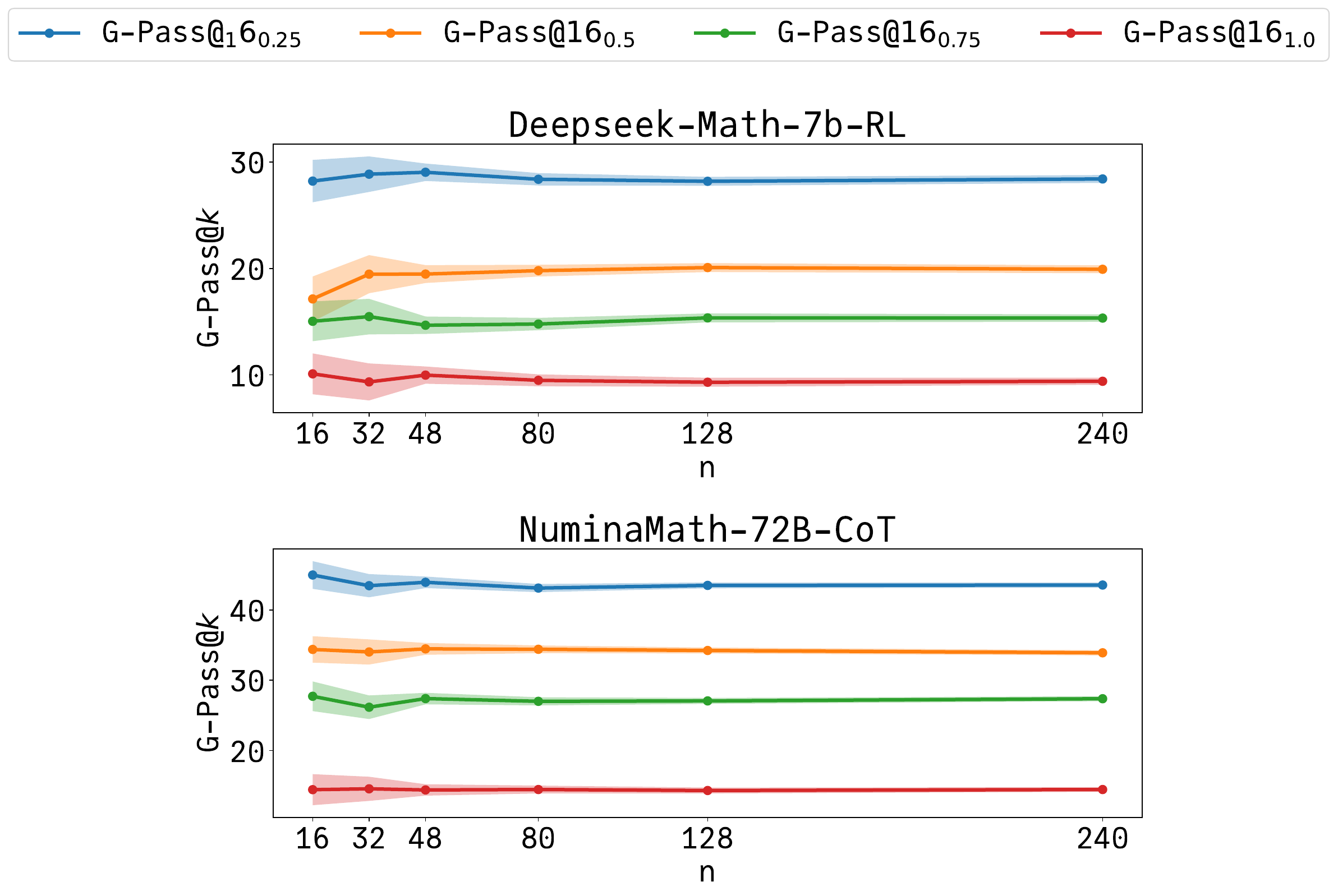}
    \caption{Illustration of \gpk \textit{w.r.t.} different values of $n$ for DeepSeek-Math-7b-RL and NuminaMath-72B-CoT on LiveMathBench.}
    \label{fig:g-pass-at-k-wrt-n}
\end{figure}

\paragraph{Performance \textit{w.r.t.} $k$. } \label{sec:performance_wrt_k}
\Cref{fig:g-pass-at-k-wrt-k} presents the results of selected models for G-Pass@$4$, G-Pass@$8$, and G-Pass@$16$. 
From the experiments, G-Pass@$k$ can achieve consistent evaluation results under different $ k$ values. 
In addition, for advanced reasoning models with strong performance, a larger value of $ k$ has better differentiation.

\paragraph{Performance \textit{w.r.t.} $n$. } \label{sec:performance_wrt_n}
As previously noted, the number of attempts $n$ is crucial for the accuracy of the estimation. 
We select two models, DeepSeek-Math-7b-RL and NuminaMath-72B-CoT, to conduct experiments with $n = \{16\} \times \{1,2,3,5,8,15\} = \{16, 32, 48, 128, 240\}$, and report G-Pass@$16_\tau$. 
The results are illustrated in \Cref{fig:g-pass-at-k-wrt-n}. 
When $n$ is small, the estimation deviation is large, as shown by the significant fluctuations in the G-Pass@$16_\tau$ values for both models. 
Conversely, for larger $n$, G-Pass@$16_\tau$ tends to stabilize, indicating a more consistent and reliable performance. 
Empirically, we recommend making at least $n = 3k$ generations when calculating \gpk to ensure estimation accuracy. 

\subsection{Impact of Sampling Parameters} \label{sec:performance_wrt_sampling}
The configuration of sampling parameters influences the diversity of model generation and, consequently, its reasoning stability. 
To investigate the effects of various sampling parameters, we evaluate models across different combinations of these parameters. 
Specifically, we focus on 3 key parameters: \emph{temperature}, \emph{top-p}, and \emph{top-k}. 
For each evaluation, we adjust one parameter while holding the others constant. 
\Cref{fig:g-pass-at-k-wrt-sampling} illustrates the experimental results for the Llama-3.3-70B-Instruct, Qwen2.5-72B-Instruct, and QwQ-32B-Preview models. 
Additional details can be found in \Cref{app:full_performance_temp}, \Cref{app:full_performance_topp}, and \Cref{app:full_performance_topk}. 
Our findings can be summarized as follows:

\begin{figure}[t]
    \centering
    \includegraphics[width=\linewidth]{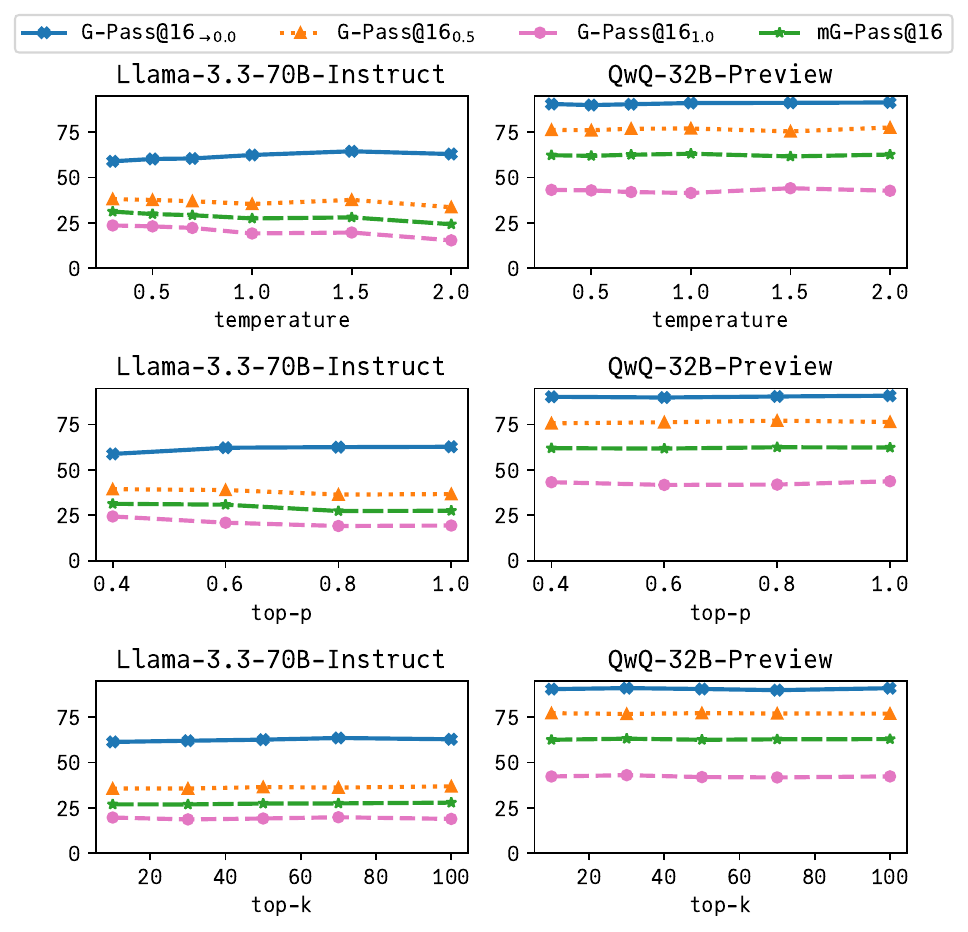}
    \caption{\gpkt performance of LLMs \textit{w.r.t.} sampling parameters, \textit{i.e.}, top-p, top-k, and tempreature.}
    \label{fig:g-pass-at-k-wrt-sampling}
\end{figure}

\paragraph{1) \gpkt Demonstrates Robustness Across Sampling Parameters. }
The proposed metric, \gpkt, demonstrates stability across various models under different sampling parameters. 
This indicates that \gpkt serves as a robust metric, accurately reflecting model performance.

\paragraph{2) Sensitivity of Different Models to Sampling Parameters Varies Greatly. }
Experimental results show that some models, including Llama-3.3-70B-Instruct, Mistral-Large-Instruct-2411, and Qwen2.5-72B-Instruct, maintain stable performance across various sampling parameters. 
We hypothesize that models with more parameters exhibit greater resistance to perturbations after adequate pre-training. 
In contrast, certain models, such as Qwen2.5-Math-72B-Instruct, experience significant performance degradation under atypical sampling parameters. 
This degradation may be attributed to the effects of post-training, making these models more susceptible to perturbations.

\paragraph{3) Long CoT Can Help to Enhance the Robustness of the Model. }
An intriguing observation is that several o1-like models exhibit considerable robustness to sampled parameters. 
For instance, the performance of QwQ remains consistently stable across all experiments. 
We believe that long COT reasoning aids the model in rectifying errors introduced by random sampling. 
This finding underscores the importance of backtracking and self-reflection in addressing complex reasoning problems.

\subsection{Impact of Question Difficulty} \label{sec:performance_wrt_difficulty}

We also examine the performance of models with respect to questions of varying difficulty levels. 
We analyze CCEE and WLPMC splits from LiveMathBench. 
CCEE is a college entrance examination that primarily involves fundamental high school mathematics knowledge, whereas WLPMC is a prestigious collegiate mathematics competition that presents significantly greater challenges.
\Cref{tab:difficulty_performance} shows the experimental results. 
The findings indicate that models struggle more with generalizing this ability to challenging questions.
The sharp decline in performance on WLPMC, compared to CCEE, suggests a higher level of difficulty models face when grasping these questions. 
For instance, on WLPMC, the Greedy Accuracy of QwQ-32B-Preview declines by $69.1$\% compared to its Pass@$16$ performance.
In contrast, on CCEE, the Greedy Accuracy of QwQ-32B-Preview only declines by $6.3$\% compared to its Pass@$16$ performance.

From the observation, we conjecture that models tend to learn superficial patterns from training data, which is reflected in the marked improvement of the Pass@$k$ metric. 
However, this increase does not necessarily translate into an enhancement of the model’s real reasoning capabilities. 

\section{Enhancing the Stability of Reasoning}

In this section, we aim to explore factors that can improve the reasoning stability. 
Initially, we assess whether straightforward supervised fine-tuning on specific corpora enhances reasoning stability. 
Subsequently, we explore the reasons for the pronounced stability of o1-like models, positing that the characteristics of their reasoning paths contribute to this improved stability.

\subsection{Can SFT Enhance Model Stability? An Empirical Analysis of Data Contamination} \label{sec:performance_wrt_contamination}

Currently, most large language models are trained on massive pre-training corpora which may lead to data contamination. 
Data contamination arises when the test data is mixed into training data, also referred to as data leakage~\citep{Why_data_contamination_issue, dong2024generalizationmemorizationdatacontamination}. 
Additionally, supervised fine-tuning on specific datasets is a common post-training method to enhance model performance. 
In this section, we explore whether training a model on a particular corpus significantly affects its reasoning stability. 

\begin{figure}[ht]
    \centering
    \includegraphics[width=.9\linewidth]{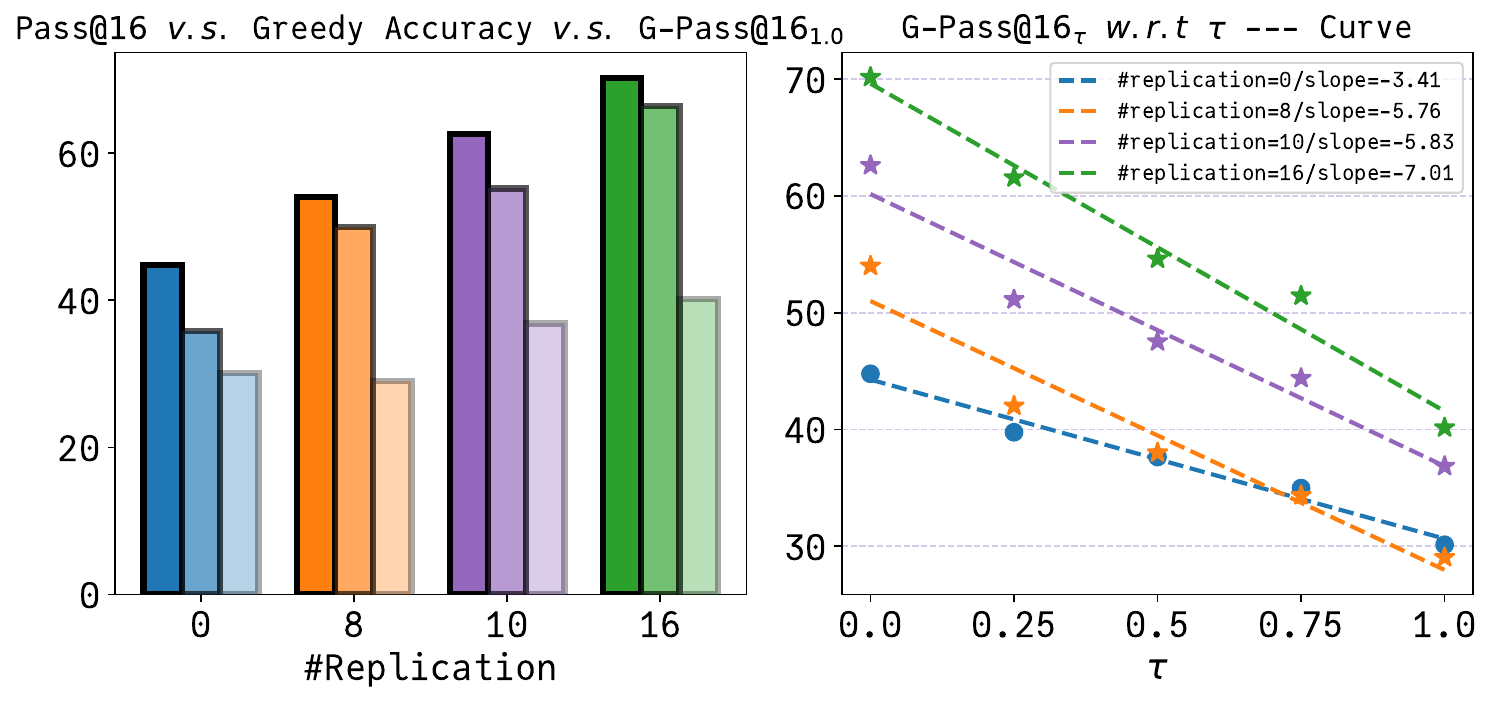}
    \caption{The data contamination experiment involves different contamination rounds, where $\#Replication$ represents the number of these rounds. The term $Slope$ denotes the slope value of the G-Pass@$16_\tau$ curve with respect to $\tau$.}
    \label{fig:data-contamination}
\end{figure}

To investigate the effect of SFT, we train the Qwen2.5-7B model on a previously unseen corpus and evaluate its reasoning performance and stability throughout the training process to ascertain whether mere SFT~(which we also refer to as overfitting or contamination) significantly impacts reasoning stability.
To be specific, the training process begins with a base set of 200,000 randomly sampled instructions from the Numina-Math-CoT corpus~\citep{numina_math_datasets}, which serves as the uncontaminated training set. 
Subsequently, we introduce incremental rounds of data contamination, consisting of 0, 6, 8, 10, and 16 rounds, where a round of 0 indicates the absence of contamination, i.e., training exclusively on the original NuminaMath data. 
The model's efficacy was assessed across these five conditions, as illustrated in \Cref{fig:data-contamination}.

Despite the observed increase in greedy score with escalating rounds of contamination, the stability, as quantified by the \gpk metric, did not exhibit a corresponding enhancement. 
Specifically, in \Cref{fig:data-contamination} left part, the disparity between actual performance (Greedy Accuracy) and stability across multiple samples (\(\gpk@16_{\tau=1.0}\)) at each contamination round—6, 8, 10, and 16—was 22, 20, 18, and 26, respectively. In contrast, this gap for a non-contaminated model was only 5, which indicates that the discrepancy between performance and stability in contaminated models is more than three times greater. 
Additionally, as shown in \Cref{fig:data-contamination} right part, the slope becomes increasingly steep as the rounds of contamination increase, showing a deterioration in model stability with each additional round of contamination. 
This phenomenon is particularly significant in certain downstream training scenarios where overfitting becomes necessary, such as in contexts characterized by data scarcity. 
In these cases, while an increase in greedy accuracy might be achieved, it often comes at the cost of reduced stability. Notably, the aforementioned performance-stability gap may not necessarily narrow proportionally with increased levels of contamination.

From the experimental results, we can contend that simple SFT does not significantly enhance the model's reasoning stability, or that LLMs are more inclined to memorize during SFT, rather than improving genuine reasoning ability~\citep{chu2025sftmemorizesrlgeneralizes}.

\subsection{Mechanistic Analysis of Long CoT in Enhancing Model Stability} \label{sec:long_cot_improve}

From the experimental results presented in \Cref{tab:main_performance} and \Cref{tab:public_performance}, it is evident that o1-like LLMs demonstrate greater stability when compared to general LLMs (e.g., Llama-3.1-8B-Instruct \& Skywork-o1, and Qwen2.5-32B-Instruct \& QwQ-32B-Preview). 
In this section, we explore the reasons behind the superior inference performance and stability of o1-like LLMs.

\begin{figure}[htb]
    \centering
    \includegraphics[width=\linewidth]{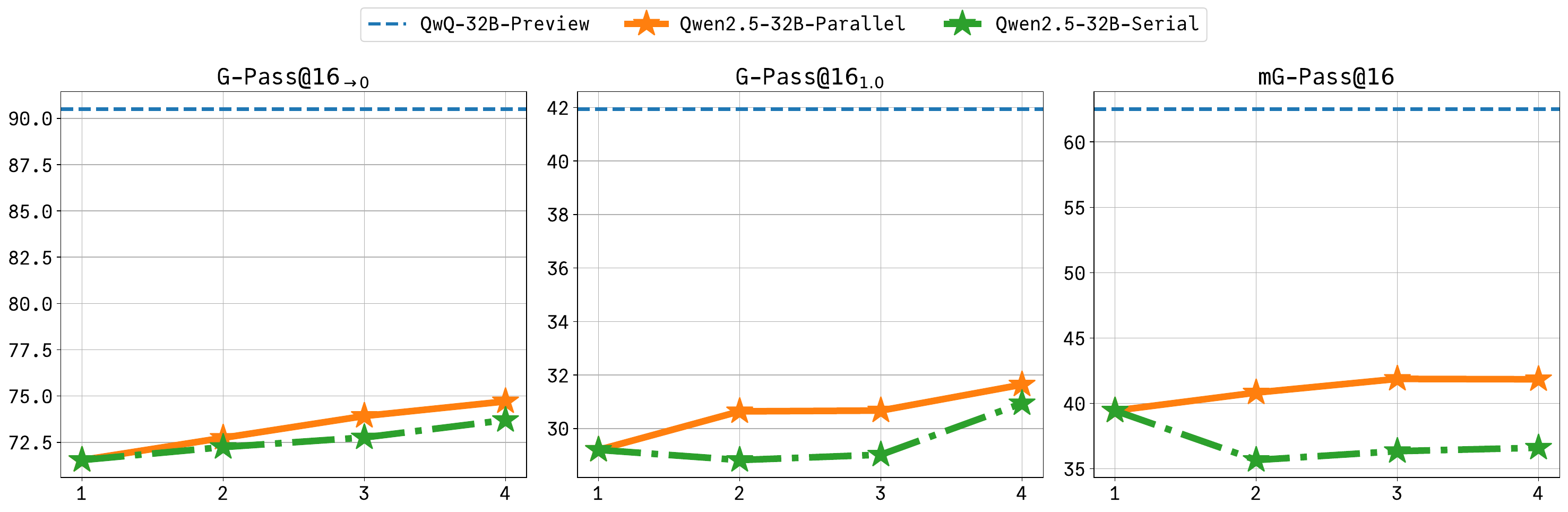}
    \caption{Comparison between Qwen2.5-32B-Instruct and QwQ-32B-Preview.}
    \label{fig:general_v-s_o1}
\end{figure}

We select Qwen2.5-32B-Instruct and QwQ-32B-Preview as the foundational models. 
Our first observation is that o1-like LLMs significantly surpass general LLMs in terms of the length of reasoning paths. 
This leads us to hypothesize whether extending the reasoning path length of general LLMs can enhance their reasoning performance and stability. 
To verify this hypothesis, we designed the following experiments: 
1) leverage Qwen2.5-32B-Instruct to perform parallel reasoning $m$ times, treating $m$ reasoning paths combined as a single reasoning process; and 
2) leverage Qwen2.5-32B-Instruct to execute serial reasoning $m$ times, inserting a break-link (e.g., \textit{Wait, let's re-evaluate the solution steps.}) between each reasoning path. 
We subsequently compared the performance of these two enhanced reasoning models with that of QwQ-32B-Preview, and the results are illustrated in \Cref{fig:general_v-s_o1}.
We observe that: 
1) while parallel reasoning in general LLMs, combined with majority voting, can improve reasoning performance and stability to some extent, this improvement is limited. 
Such performance and stability remain significantly inferior to those of o1-like LLMs when considering an equivalent number of reasoning tokens at the same scale. 
While parallel reasoning is anticipated to outperform certain models with more parameters as the number of tokens rises, it unavoidably incurs substantial reasoning overhead. 
2) a robust backbone model possesses the capability to reflect and backtrack. Introducing specialized break-links can enhance this capability to some extent, thereby improving the performance and stability of reasoning. However, this capability requires further stimulation compared to models similar to o1-like LLMs.

Therefore, we propose a preliminary hypothesis: \textbf{o1-like LLMs exhibit superior reasoning performance and stability due to their reflective and backtracking capabilities, which enable them to correct errors in the reasoning path}. 
The reasoning path can be conceptualized as a gradient descent process in which general LLMs may easily become trapped in local optima when optimizing in a single direction. 
In contrast, o1-like LLMs, through their backtracking and reflection abilities, can escape local optima and more effectively identify global optima.

To validate our hypothesis, we investigate the hidden states of LLMs during the reasoning process, which reflects the decision process of LLMs~\citep{DBLP:conf/iclr/0006LZKSLL23,DBLP:journals/corr/abs-2410-13640}. 
Specifically, we utilize the CoE-Score to quantitatively analyze the variation tendencies of hidden states across different layers of LLMs. 
For detailed information, please refer to the original paper by \citet{DBLP:journals/corr/abs-2410-13640}. 
We choose several questions from LiveMathBench and prompted Qwen2.5-32B-Instruct and QwQ-32B-Preview to perform multiple reasoning. Subsequently, we gather all candidate answers generated by Qwen2.5-32B-Instruct and QwQ-32B-Preview and calculate the CoE-Scores with respect to candidate answers at each position of the reasoning paths.
The curves are depicted in \Cref{fig:coe_score1}, \Cref{fig:coe_score2}, \Cref{fig:coe_score3}, and \Cref{fig:coe_score4}. 
These figures reveal that o1-like LLMs exhibit multiple decision changes during the reasoning process, as indicated by frequent peaks and troughs in their CoE-Score curves. 
In contrast, the decision process of general LLMs is more linear, resulting in a notably flat CoE-Score curve. 
This linearity makes general LLMs more susceptible to noise and incorrect intermediate steps, thereby reducing their inference performance and stability.

\section{Related Work}

\paragraph{Stability of LLM Reasoning. }
Large language models (LLMs) exhibit remarkable performance in reasoning tasks, encompassing question answering, programming, and mathematical problem-solving.
Despite their prowess, the output stability of LLMs poses a significant challenge, whereby the model's outputs can vary for the same input due to random sampling or hallucinations, impacting the model's reliability and predictability~\citep{DBLP:journals/corr/abs-2406-19470,DBLP:journals/corr/abs-2406-16203,DBLP:journals/corr/abs-2408-04667,DBLP:conf/emnlp/ZhuoZFDL024}.
\citet{DBLP:journals/corr/abs-2408-04667} introduced two new metrics: TARr@N for the total agreement rate at N runs over raw output and TARa@N for total agreement over parsed-out answers.
However, TARr@N and TARa@N focus solely on measuring output consistency, our work introduces a novel evaluation metric \gpk for evaluating the mathematical reasoning proficiency of LLMs. 
This metric aims to assess the model's true reasoning ability by not only considering output consistency but also emphasizing correctness.
Additionally, \citet{DBLP:journals/corr/abs-2406-12045} propose pass\^{}k, an approach to assess the reliability and consistency of real-world agent tasks, sharing the same formulation as \gpk. We relax the overly restrictive conditions of \gpk and further introduce \gpkt and m\gpk. In comparison to the original \gpk and pass\^{}k, our methods offer enhanced flexibility and more comprehensive evaluation capabilities.

\paragraph{Mathematical Reasoning Benchmarks for LLMs. }
The assessment of large language models (LLMs) in mathematical reasoning has led to the development of specialized benchmarks focusing on different aspects of an LLM's mathematical proficiency.
GSM8K~\citep{DBLP:journals/corr/abs-2110-14168} presents a dataset composed of elementary-level math word problems, segregated into training and testing sets, that demand multi-step reasoning and detailed solution paths.
MATH~\citep{DBLP:conf/nips/HendrycksBKABTS21} encompasses 12,500 problems derived from high school math competitions, challenging LLMs with advanced topics like calculus and algebra, and providing step-by-step solutions to facilitate coherent training.
MathBench~\citep{DBLP:conf/acl/LiuZQDFZZZLC24} is a hierarchical benchmark that assesses both theoretical and applied mathematical abilities, consisting of 3,709 questions spanning basic arithmetic to university level, structured across five educational tiers.
Omni-Math~\citep{DBLP:journals/corr/abs-2410-07985} focuses on Olympic-level mathematical reasoning, featuring 4,428 competition-level problems categorized into over 33 subfields and 10 difficulty levels, spanning from entry-level to professional international competitions.
\section{Conclusion}

In this work, we propose \gpkt and mG-Pass@$k$, novel evaluation metrics that assess both the reasoning capability and performance consistency of LLMs across varying correctness thresholds. 
Through detailed evaluations conducted on mathematical reasoning benchmarks, we find that current LLMs struggle with consistent reasoning. 
Additionally, we demonstrate \gpk's robustness and preliminarily explore strategies for enhancing the models' reasoning stability.

\section*{Limitations}
In this study, we analyze the stability of large language models in reasoning tasks, propose a new metric, and conduct experiments on various mathematical reasoning benchmarks. Due to constraints in space and resources, our work has the following limitations:
\begin{itemize}[leftmargin=*]
    \item our experiments are not extended to broader reasoning tasks. Nevertheless, we assert that mathematical reasoning, as a representative task, aptly supports our experiments and conclusions. Future research will address a wider range of reasoning tasks.
    \item We attempted to run experiments on as many large language models (LLMs) as possible; however, due to hardware limitations, API call overhead, and the service instability of some APIs, a few experimental results were missing. However, we believe that the experimental results presented in this paper are sufficient to support our conclusions. 
    \item The benchmarks in this paper include only English and Chinese, however, the conclusions are generally applicable to large language models of all languages.
\end{itemize}

\section*{Acknowledgements}
This work was supported by National Key R\&D Program of
China 2022ZD0161600, and Shanghai Oriental Talents Project BJZH2024070.

\bibliography{refs}

\clearpage
\newpage
\appendix
\section{Detailed Related Work}

\section{Estimation of \texorpdfstring{\gpk}{G-Pass@k}} \label{app:k_at_pass_estimation}

\begin{figure*}[!ht]
    \centering
    \includegraphics[width=.9\linewidth]{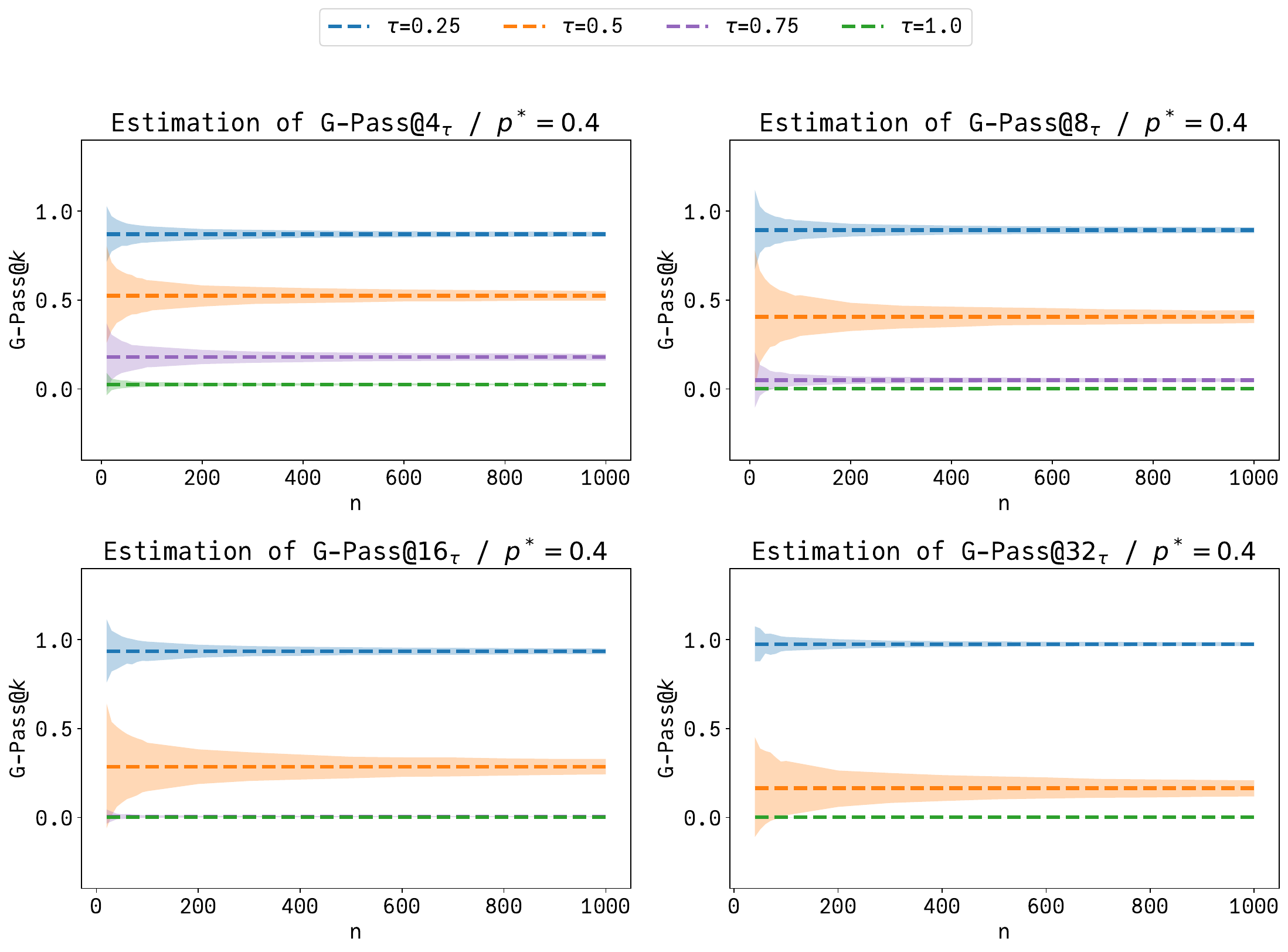}
    \caption{Illustration of estimation and the true value of \gpkt.}
    \label{fig:est-k-pass}
\end{figure*}

To demonstrate the unbiasedness of \Cref{eq:k_at_pass}, we conduct the simulation experiment illustrated in \Cref{fig:est-k-pass}. 
Specifically, we assume the probability of a model providing the correct solution in a single run is $p^* = 0.4$. 
For each $n$, we perform several random Bernoulli samplings to obtain different values of $c$ to calculate \gpkt, and then compute the mean and variance to generate the figure. 
From \Cref{fig:est-k-pass}, it can be observed that \Cref{eq:k_at_pass} is an unbiased estimator, facilitating fair comparison across different values of $n$.

\section{Proof of Theorem 2.1} \label{app:prof_k_at_pass_equal_pass_at_k}

\begin{proof}

    Since $j$ starts iterating at the upward rounding of $\lceil \tau \cdot k \rceil$ and $\tau \in (0, 1]$, so we have:
    
    \begin{equation} \label{eq:proof_1}
        \lim_{\tau \rightarrow 0} \sum_{j= \lceil \tau \cdot k \rceil}^{c} \frac{{c \choose j} \cdot {{n - c} \choose {k - j}}}{{n \choose k}} =  \sum_{j=1}^{c} \frac{{c \choose j} \cdot {{n - c} \choose {k - j}}}{{n \choose k}}.
    \end{equation}

    According to the Vandermonde's Identity~\citep{vandermonde1772}, the numerator term on the right side of \Cref{eq:proof_1} can be written as

    \begin{equation}
        \begin{aligned}
            & \;\;\;\; \sum_{j=1}^{c} {c \choose j} \cdot {{n - c} \choose {k - j}} \\
            &= \sum_{j=0}^{c} {c \choose j} \cdot {{n - c} \choose {k - j}} - {{n - c} \choose k} \\
            &= {n \choose k} - {{n - c} \choose k}.
        \end{aligned}
    \end{equation}

    So we conclude that:

    \begin{equation}
        \begin{aligned}
            \lim_{\tau \rightarrow 0} \sum_{j= \lceil \tau \cdot k \rceil}^{c} \frac{{c \choose j} \cdot {{n - c} \choose {k - j}}}{{n \choose k}} & =  \sum_{j=1}^{c} \frac{{c \choose j} \cdot {{n - c} \choose {k - j}}}{{n \choose k}} \\
            & = \frac{{n \choose k} - {{n - c} \choose k}}{{n \choose k}} \\
            & = 1 - \frac{{{n - c} \choose k}}{{n \choose k}}.
        \end{aligned}
    \end{equation}
    
\end{proof}

\section{\texorpdfstring{Pass@$k$}{Pass@k} \textit{v.s.} \texorpdfstring{\gpk}{G-Pass@k}}

\begin{figure*}[t]
    \centering
    \includegraphics[width=.9\linewidth]{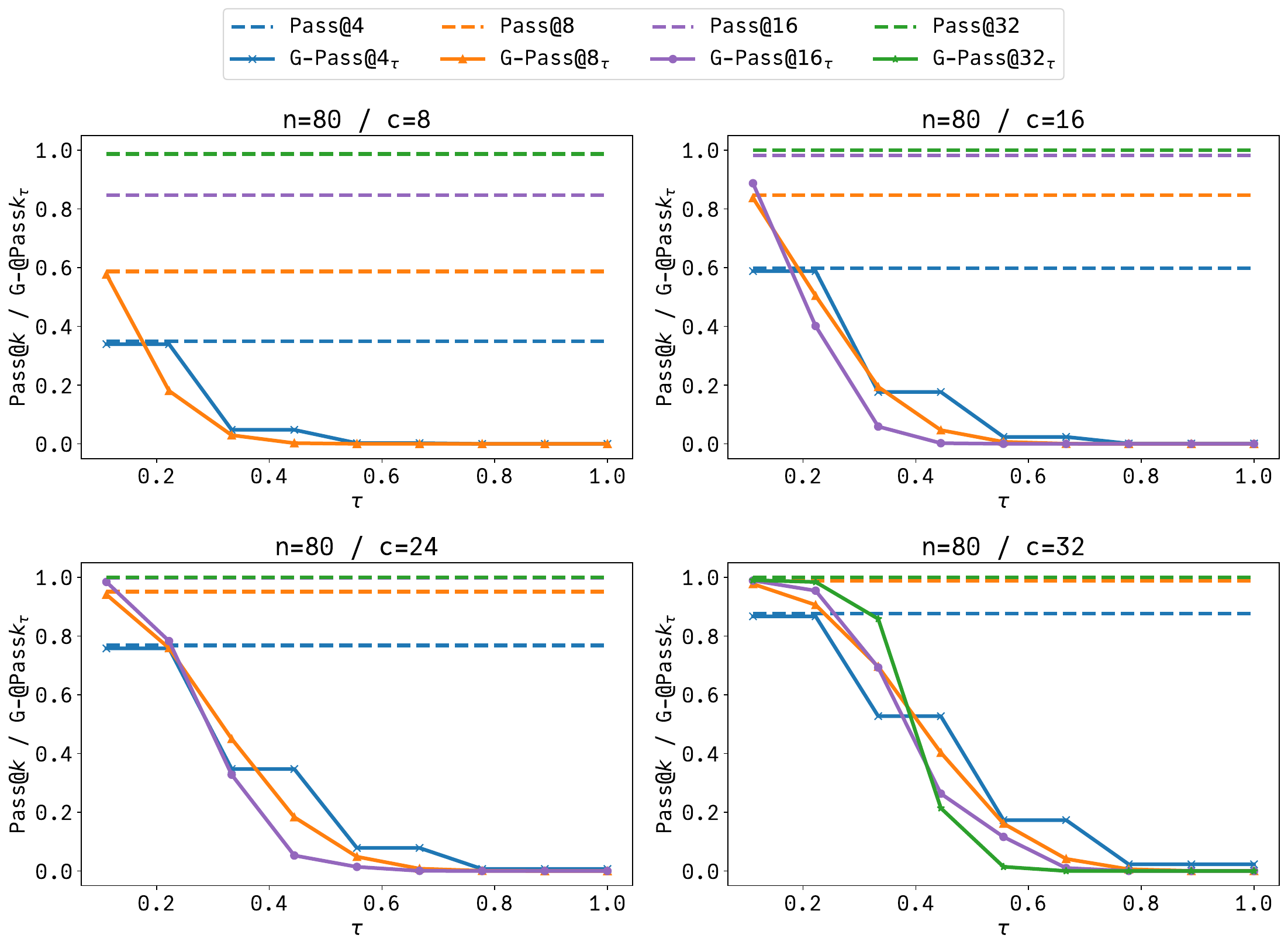}
    \caption{Comparison of Pass@$k$ and \gpk. In our simulation setup, we set $n=10$ and $c=\{8, 16, 24, 32\}$, and then calculate Pass@$k$ and G-Pass@$k$.}
    \label{fig:pass-k-vs-k-pass}
\end{figure*}

To facilitate an intuitive comparison between Pass@$k$ and \gpk, \Cref{fig:pass-k-vs-k-pass} presents the metric values for various $c$ values with $n=80$. The figure illustrates that, while Pass@$k$ offers insights into the model's capabilities, relying solely on it may lead to an overestimation of the model's actual performance. For instance, as shown in the upper left of \Cref{fig:pass-k-vs-k-pass}, even when the model solves the question correctly only 8 times out of 80 runs, Pass@$k$ produces a notably high score (Pass@$k$ $> 0.8$ for $k \geq 16$). Additionally, as $c$ increases beyond a certain threshold, differentiating performance based on Pass@$k$ becomes increasingly challenging.

In contrast, \gpk provides a more accurate and nuanced evaluation of the model's performance, as depicted in \Cref{fig:pass-k-vs-k-pass}. Across varying $c$ values, \gpk demonstrates clear distinctions between performance levels. 
Moreover, by adjusting the threshold parameter $\tau$, \gpk can highlight different aspects of the model's performance: a lower threshold emphasizes the model's potential, while a higher threshold underscores its stability and mastery of the problem.

In summary, \gpk not only delivers a more comprehensive performance assessment compared to Pass@$k$, but also, through flexible threshold configurations, effectively balances the evaluation of the model's potential and stability.

\section{Evaluated LLMs} \label{app:evaluated_llm}
In this paper, we conduct experiments on LLMs including:
InternLM3-8B-Instruct~\citep{DBLP:journals/corr/abs-2402-06332,internlm8b}, 
DeepSeek-Math-7b-RL~\citep{DBLP:journals/corr/abs-2402-03300}, 
DeepSeek-V3.0-Chat~\citep{DBLP:journals/corr/abs-2412-19437}, 
Gemma-2-27b-it~\citep{DBLP:journals/corr/abs-2408-00118}, 
Llama-3.1-8B-Instruct~\citep{DBLP:journals/corr/abs-2407-21783}, 
Llama-3.1-70B-Instruct~\citep{DBLP:journals/corr/abs-2407-21783}, 
Yi-1.5-34B-Chat~\citep{DBLP:journals/corr/abs-2403-04652}, 
Llama-3.3-70B-Instruct~\citep{DBLP:journals/corr/abs-2407-21783,llama3_70b_instruct}, 
NuminaMath-72B-CoT~\citep{numina_math_7b},  
Mistral-Large-Instruct-2411~\citep{mistral_large_instruct_2411}, 
Qwen2.5-7B-Instruct~\citep{DBLP:journals/corr/abs-2412-15115}, 
Qwen2.5-Math-7B-Instruct~\citep{DBLP:journals/corr/abs-2409-12122}, 
Qwen2.5-32B-Instruct~\citep{DBLP:journals/corr/abs-2412-15115}, 
Qwen2.5-72B-Instruct~\citep{DBLP:journals/corr/abs-2412-15115}, 
Qwen2.5-Math-72B-Instruct~\citep{DBLP:journals/corr/abs-2409-12122}, 
Qwen2.5-Max~\citep{DBLP:journals/corr/abs-2412-15115}, 
Claude-3.5-Sonnet~\citep{claude35sonnet}, 
Gemini-1.5-Pro~\citep{gemini15pro}, 
Gemini-2.0-Flash-Exp~\citep{gemini20exp}, 
and GPT-4o-2024-11-20~\citep{gpt4o}. 
Additionally, we include several o1-like LLMs, such as 
QwQ-32B-Preview~\citep{qwq-32b-preview}, 
Skywork-o1-Open-Llama-3.1-8B~\citep{skyworkopeno12024}, 
DeepSeek-R1 series models~\citep{deepseekai2025deepseekr1incentivizingreasoningcapability}, 
OpenAI o1-mini~\citep{o1preview}, 
and OpenAI o3-mini~\citep{o3mini}.

\section{Data Details} \label{app:data_details}

\subsection{LiveMathBench}

To effectively analyze the reasoning stability of large language models, we construct a new benchmark named LiveMathBench. 
LiveMathBench consists of the latest complex mathematical questions from various examinations and competitions, ensuring minimal risk of data contamination. 
LiveMathBench will undergo ongoing updates with new questions to continuously evaluate the mathematical reasoning performance of models.

\subsubsection{Benchmark Construction}
LiveMathBench is specifically designed to include out-of-domain question sets with different difficulty spans from various mathematical exams and competitions, aiming to avoid data contamination issues in existing LLMs and public math benchmarks~\citep{DBLP:journals/corr/abs-2311-01964,DBLP:conf/emnlp/0001GGL24,DBLP:journals/corr/abs-2409-01790}.
LiveMathBench~(version of 202412) incorporates the latest problems from the China National Mathematical Olympiad (CNMO), China's College Entrance Examination (CCEE), American Mathematics Competition (AMC), and William Lowell Putnam Mathematical Competition (WLPMC). 
These datasets encompass diverse levels of difficulty and linguistic variations and have low overlap with publicly available datasets, ensuring a comprehensive evaluation of the generalization capabilities of LLMs across various mathematical scenarios. 

\begin{table*}[!ht]
    \centering
    \caption{Statistics of LiveMathBench}
    \label{tab:benchmark_statistics}
    \begin{tabular}{lcccc}
        \toprule
        \textbf{Dataset} & \textbf{Language} & \textbf{\#Fill-In-the-Blank} & \textbf{\#Problem-Solving} & \textbf{\#Questions}\\ 
        \midrule
        CNMO & en \& cn & - & 18$_{\times2}$ & 18$_{\times2}$ \\ 
        CCEE & en \& cn & 13$_{\times2}$ & 31$_{\times2}$ & 44$_{\times2}$ \\ 
        AMC & en \& cn & - & 46$_{\times2}$ & 46$_{\times2}$ \\ 
        WLPMC & en \& cn & - & 11$_{\times2}$ & 11$_{\times2}$ \\ 
        \midrule
        ALL & en \& cn & 13$_{\times2}$ & 106$_{\times2}$ & 119$_{\times2}$ \\ 
        \bottomrule
    \end{tabular}
\end{table*}

\subsubsection{Data Sources}

LiveMathBench~(version of 202412) is composed of 4 parts including CNMO, CCEE, AMC, and WLPMC.

\paragraph{CNMO. }
The CNMO section features curated questions from the latest \underline{\textbf{C}}hinese \underline{\textbf{N}}ational \underline{\textbf{M}}athematics \underline{\textbf{O}}lympiad. 
To enhance the difficulty level, single-choice questions are transformed into problem-solving tasks by concealing answer options, necessitating models to reason independently and provide solutions.

\paragraph{CCEE. }
In the CCEE segment, we have selected questions from recent mock exams of \underline{\textbf{C}}hina's \underline{\textbf{C}}ollege \underline{\textbf{E}}ntrance \underline{\textbf{E}}xamination, excluding multi-modal proof problems. 
We have excluded multiple-choice questions and converted single-choice items into problem-solving questions, removing provided answer choices to assess the models' ability to generate solutions autonomously.

\paragraph{AMC. }
The AMC section includes questions from the latest \underline{\textbf{A}}merican \underline{\textbf{M}}athematics \underline{\textbf{C}}ompetition, where each original question typically offers five possible answers labeled A through E, with only one correct option. 
Consistent with our approach in other sections, we convert these single-choice questions into problem-solving cues, encouraging models to deduce solutions without the aid of provided options.

\paragraph{WLPMC. }
We also include questions from the latest \underline{\textbf{W}}illiam \underline{\textbf{L}}owell \underline{\textbf{P}}utnam \underline{\textbf{M}}athematical \underline{\textbf{C}}ompetition (WLPMC). 
Regarded as one of the most prestigious university-level mathematics competitions globally, the WLPMC challenges participants with problems that span a broad spectrum of mathematical disciplines. 
These include geometry, algebra, trigonometry, calculus, linear algebra, combinatorics, probability theory, number theory, complex numbers, and differential equations. 

\subsubsection{Benchmark Statistics}
\Cref{tab:benchmark_statistics} presents comprehensive statistics for the LiveMathBench. 
In order to enhance benchmark diversity and assess the performance of LLMs in multilingual settings, both English and Chinese versions of the questions are included.

\subsubsection{Data Samples}

Here we provide some samples in LiveMathBench.
\begin{figure}[!ht] 
    \begin{AIbox}{Example in CNMO}
        [Question]\par
        \begin{CJK*}{UTF8}{gbsn}
            设复数$z,w$满足$z+w=2$，求$S=|z^2-2w|+|w^2-2z|$的最小可能值。
        \end{CJK*} \par
        [Answer] \par
        $8\sqrt{5} - 16$ \par
        [Question Type] \par
        \begin{CJK*}{UTF8}{gbsn}
            问答
        \end{CJK*}
    \end{AIbox} 
\end{figure}

\begin{figure}[!ht] 
    \begin{AIbox}{Example in CCEE}
        [Question]\par
        \begin{CJK*}{UTF8}{gbsn}
            函数$f(x)=\frac{x^3e^{3x}-3\\ln x-1}{x}(x>0)$的最小值是\underline{\;\;\;\;\;\;\;}
        \end{CJK*} \par
        [Answer] \par
        $3$ \par
        [Question Type] \par
        \begin{CJK*}{UTF8}{gbsn}
            填空
        \end{CJK*}
    \end{AIbox} 
\end{figure}

\begin{figure}[!ht] 
    \begin{AIbox}{Example in AMC}
        [Question]\par
        The graph of $y=e^{x+1}+e^{-x}-2$ has an axis of symmetry. What is the reflection of the point $(-1,\frac{1}{2})$ over this axis? \par
        [Answer] \par
        $\left(0,\frac{1}{2}\right)$ \par
        [Question Type] \par
        Problem-Solving
    \end{AIbox} 
\end{figure}

\begin{figure}[!ht] 
    \begin{AIbox}{Example in WLPMC}
        [Question]\par
        A sequence $y_1,y_2,\dots,y_k$ of real numbers is called \emph{zigzag} if $k=1$, or if $y_2-y_1, y_3-y_2, \dots, y_k-y_{k-1}$ are nonzero and alternate in sign. Let $X_1,X_2,\dots,X_n$ be chosen independently from the uniform distribution on $[0,1]$. Let $a(X_1,X_2,\dots,X_n)$ be the largest value of $k$ for which there exists an increasing sequence of integers $i_1,i_2,\\dots,i_k$ such that $X_{i_1},X_{i_2},\dots,X_{i_k}$ is zigzag. Find the expected value of $a(X_1,X_2,\dots,X_n)$ for $n \geq 2$. \par
        [Answer] \par
        $\frac{2n+2}{3}$ \par
        [Question Type] \par
        Problem-Solving
    \end{AIbox} 
\end{figure}

\subsubsection{Hard Split}
To achieve better differentiation, we also implemented a hard split for evaluating models with strong inference performance. Specifically, we selected subsets where QwQ-32B-Preview and OpenAI o1-mini models demonstrated lower accuracy, consisting of 21 questions in English and 24 questions in Chinese.




\subsection{MATH500-L5}
MATH500~\citep{DBLP:conf/iclr/LightmanKBEBLLS24} dataset is a curated subset of a larger collection MATH~\citep{DBLP:conf/nips/HendrycksBKABTS21}, intended to challenge LLMs with complex mathematical problems. It encompasses a variety of advanced questions from multiple domains including algebra, geometry, probability, and number theory, thereby providing a comprehensive assessment of a model's proficiency in mathematical reasoning.
We select all questions with difficulty 5, resulting in \textbf{MATH500-L5}, which contains $134$ questions.

\subsection{AIME2024-45}
Tailored for evaluating LLM performance at the American Invitational Mathematics Examination~(AIME) level, the AIME question set~\citep{AIME2024,DBLP:journals/corr/abs-2409-12122} presents a series of intricate tasks that test logical thinking, abstract reasoning, and accurate calculation skills. 
This resource aims to push the boundaries of what LLMs can achieve in solving sophisticated mathematical problems.
We combine the part 1 and the part 2 of the American Invitational Mathematics Examination 2024, resulting in $45$ questions, called \textbf{AIME2024-45}.

\subsection{AIME2025}
We also incorporate Part 1 of the American Invitational Mathematics Examination 2025, referred to as \textbf{AIME2025}~\citep{AIME2025}, as our evaluation benchmark, comprising $15$ questions.

\section{Judge Details} \label{app:judge}

\subsection{Configurations of Judge Model}
Inspired by previous works~\citep{DBLP:conf/nips/ZhengC00WZL0LXZ23,DBLP:journals/corr/abs-2409-11239}, we leverage Qwen2.5-72B-Instruct~\citep{DBLP:journals/corr/abs-2407-10671} to judge if the answers generated by the models are consistent with the golden answers, consider the high inference cost of the closed source models such as OpenAI models.
We set the temperature to 0.0, and maximum output tokens to $8,192$.

\subsection{Prompt for Judge}

We leverage the prompts shown in Box \texttt{Chinese Version of Judge Prompt} and Box \texttt{English Version of Judge Prompt} to judge the consistency between candidate answers and reference answers.

\begin{figure*}
    \begin{AIbox}[label=box:judge_cn]{Chinese Version of Judge Prompt}
        \begin{CJK*}{UTF8}{gbsn}
            请你作为一个数学阅卷专家，判断下面的答案是否与标准答案一致，即考生是否回答正确。下面是一些评判标准：\par
            1. 有些答案可能包含多项内容，可能有单选题，多选题，填空题和问答题，只要答案与标准答案一致即可, 对于多选题和多个空的填空题，需要考生对应的选项或空都回答正确才算正确。\par
            2. 有些答案可能通过不同的方式表达，比如有些答案可能是一个数学表达式，有些答案可能是一个文字描述，只要表达的意思一致即可。且有些公式通过不同的方式表达，但等价，也是正确的。\par
            3. 你不需要重新计算问题答案，因为标准答案已经给出，只需要根据问题形式来判断考生的答案是否与标准答案一致，是否正确即可。\par
            \par
            请你根据上述标准，判断下面的答案是否与标准答案一致，如果一致，请在最后输出\textbackslash \textbackslash boxed\{\{yes\}\}, 否则输出\textbackslash \textbackslash boxed\{\{no\}\}, 如果难以判断，请输出\textbackslash \textbackslash boxed\{\{no\}\}. \par
            原问题：\{question\} \par
            标准答案：\{reference\_answer\} \par
            考生答案：\{candidate\_answer\} \par
            \par
            分析： \par
        \end{CJK*}
    \end{AIbox}
\end{figure*}

\begin{figure*}
    \begin{AIbox}[label=box:judge_en]{English Version of Judge Prompt}
        Please act as an expert in grading mathematics exam papers, and judge whether the following answers match the standard answers, i.e., whether the examinee answered correctly. Here are some evaluation criteria: \par
        \par
        1. Some answers may contain multiple parts, such as single-choice questions, multiple-choice questions, fill-in-the-blank questions, and problem-solving questions. As long as the answer matches the standard answer, it is considered correct. For multiple-choice questions and fill-in-the-blank questions with multiple blanks, the examinee must answer all corresponding options or blanks correctly to be considered correct. \par
        2. Some answers may be expressed in different ways; for example, some answers may be mathematical expressions, while others may be textual descriptions. As long as the meaning conveyed is consistent, it is considered correct. Additionally, some formulas may be expressed differently but are equivalent, which is also considered correct. \par
        3. You do not need to recalculate the problem answers, as the standard answers are already provided. You only need to judge whether the examinee's answer matches the standard answer based on the form of the question and whether it is correct. \par
        \par
        Please judge whether the following answer matches the standard answer according to the above criteria. If they match, output \textbackslash \textbackslash boxed\{\{yes\}\}, otherwise output \textbackslash \textbackslash boxed\{\{no\}\}. If it is difficult to judge, also output \textbackslash \textbackslash boxed\{\{no\}\}. \par
        Original Question: \{question\} \par
        Standard Answer: \{reference\_answer\} \par
        Examinee's Answer: \{candidate\_answer\} \par
        \par
        Analysis: \par
    \end{AIbox}
\end{figure*}

\subsection{Evaluation of Judge Model}

To evaluate the effectiveness of our judge model, we compared the agreement rate between Qwen2.5-72B-as-Judge and GPT4o-as-Judge~\citep{gpt4o}. 
Specifically, we randomly selected 300 samples from the generations of five different models and used the judgments from GPT4o as the ground truth. 
We then calculated the agreement rate between the judgments made by our model and those by GPT4o. 
\Cref{tab:evaluation_judge_model} presents the results, demonstrating that Qwen2.5-72B-as-Judge achieves high consistency with GPT4o-as-Judge across different models. 
These findings validate the feasibility of Qwen2.5-72B-as-Judge.

\begin{table*}[!ht]
    \centering
    \caption{Agreement rates between Qwen2.5-72B-as-Judge and GPT4o-as-judge.}
    \label{tab:evaluation_judge_model}
    \begin{tabular}{lccc}
        \toprule
        \textbf{Models Need to Judge} & \textbf{Agreement} & \textbf{Disagreement} & \textbf{Accuracy~(\%)} \\ 
        \midrule
        Deepseek-Math-7B-RL & $296$ & $4$ & $98.7$ \\ 
        Qwen2.5-32B-Instruct & $282$ & $18$ & $94.0$ \\ 
        Qwen2.5-Math-72B-Instruct & $287$ & $13$ & $95.7$ \\ 
        Mistral-Large-Instruct-2411 & $285$ & $15$ & $95.0$ \\ 
        QwQ-32B-Preview & $290$ & $10$ & $96.7$ \\ 
        \bottomrule
    \end{tabular}
\end{table*}

\section{Implementation Details}

In all experiments, we set the number of generations, $n$, to $16\times3 = 48$ and report the greedy accuracy, Pass@$k$~(G-Pass@$k_{\rightarrow 0}$), and \gpk values, where $k \in \{4,8,16\}$ and $\tau \in \{0.25, 0.5, 0.75, 1.0\}$. 
For the sampling parameters of open-source models, we configure the $\text{temperature}$ to $1.0$, $\text{top-p}$ to $0.8$, $\text{top-k}$ to $50$, and $\text{repetition-penalty}$ to $1.0$. 
For open-source models, the maximum number of tokens is set to $8,192$ for non-o1 LLMs and $32,768$ for o1-like LLMs. 
For closed-source models, due to constraints of inference costs, we configured the maximum completion tokens to $4,096$ for non-o1 LLMs and $8,192$ for o1-like LLMs.
We use the OpenCompass~\citep{2023opencompass} platform to evaluate all LLMs. 

Due to the diverse formats of the final answers produced by models in complex mathematical questions, we leverage Qwen-2.5-72B-Instruct~\citep{DBLP:journals/corr/abs-2407-10671} to judge whether the content generated by the tested model aligns with the standard answer. 
In our judge pipeline, we provide the original question, reference answer, and model-generated answer, prompting Qwen-2.5-72B-Instruct to determine whether the candidate solution is consistent with the reference answer.
The details of the judging process can be found in \Cref{app:judge}.

All experiments were conducted on clusters equipped with NVIDIA A800 GPUs and Intel(R) Xeon(R) Platinum 8336C CPUs, requiring hundreds of GPU hours to complete.

\section{Additional Experimental Results} \label{app:additional_experimental_results}

\subsection{Full Performance on LiveMathBench} \label{app:full_performance_livemath}

\Cref{tab:full_main_performance} presents the comprehensive performance results on LiveMathBench.

\subsection{Full Performance on LiveMathBench Hard Split} \label{app:full_performance_livemath_hard}

\Cref{tab:full_main_performance_hard} presents the comprehensive performance results on LiveMathBench hard split.

\begin{table*}[t]
    \centering
    \caption{Full performance of models on LiveMathBench. We report results of greedy decoding, Pass@$16$~(G-Pass@$16_{\rightarrow 0}$), G-Pass@$16_{\{0.25, 0.5, 0.75, 1.0\}}$, and mG-Pass@$16$.}
    \label{tab:full_main_performance}
    \resizebox{\textwidth}{!}{
        \begin{threeparttable}[b]
        \begin{tabular}{lccccccc}
            \toprule
            \multirow{2}{*}{\textbf{LLMs}} & \multirow{2}{*}{\textbf{Greedy}} & \multicolumn{6}{c}{\textbf{G-Pass@$16$}~(\Cref{eq:k_at_pass}) / \%} \\
            \cmidrule(lr){3-8}
            & & \textbf{G-Pass@$16_{\rightarrow 0}$} & \textbf{G-Pass@$16_{0.25}$} & \textbf{G-Pass@$16_{0.5}$} & \textbf{G-Pass@$16_{0.75}$} & \textbf{G-Pass@$16_{1.0}$} &  \textbf{mG-Pass@$16$} \\
            \midrule
            \multicolumn{8}{c}{\textbf{General LLMs}} \\
            \midrule
            {\small Llama-3.1-8B-Instruct} & $24.0$ & $53.3$ & $29.9$ & $18.2$ & $11.3$ & $4.5$ & $10.4$ \\
            {\small Yi-1.5-34B-Chat} & $24.8$ & $58.7$ & $31.4$ & $18.6$ & $11.3$ & $6.0$ & $11.0$ \\
            {\small Gemma-2-27b-it} & $26.9$ & $54.3$ & $33.6$ & $23.5$ & $17.8$ & $12.7$ & $17.3$ \\ 
            {\small Llama-3.1-70B-Instruct} & $29.8$ & $59.2$ & $38.6$ & $30.0$ & $22.2$ & $12.5$ & $20.8$ \\
            {\small Qwen2.5-7B-Instruct} & $37.0$ & $66.5$ & $47.3$ & $36.5$ & $27.2$ & $16.0$ & $25.8$ \\
            {\small Llama-3.3-70B-Instruct} & $40.3$ & $62.0$ & $45.8$ & $36.2$ & $28.9$ & $19.1$ & $27.5$ \\
            {\small GPT-4o-2024-11-20} \tnote{\dag} & $44.8$ & $70.8$ & $54.6$ & $41.9$ & $32.9$ & $22.2$ & $31.6$ \\
            {\small InternLM3-8B-Instruct} & $44.5$ & $69.5$ & $50.8$ & $43.0$ & $35.4$ & $23.0$ & $33.6$ \\ 
            {\small Claude-3.5-Sonnet} \tnote{\dag} & $46.7$ & $71.2$ & $54.3$ & $44.1$ & $36.2$ & $26.6$ & $35.3$ \\
            {\small Mistral-Large-Instruct-2411} & $41.6$ & $47.3$ & $42.2$ & $39.4$ & $37.1$ & $32.9$ & $36.4$ \\
            {\small Qwen2.5-72B-Instruct} & $51.7$ & $69.6$ & $55.6$ & $47.3$ & $39.6$ & $29.0$ & $37.8$ \\
            {\small Qwen2.5-32B-Instruct} & $50.8$ & $72.0$ & $57.6$ & $48.3$ & $39.5$ & $28.6$ & $38.1$ \\
            {\small Qwen2.5-Max} \tnote{\dag} & $52.9$ & $74.8$ & $62.8$ & $52.7$ & $44.3$ & $31.1$ & $42.2$ \\
            {\small Gemini-1.5-Pro-Latest} \tnote{\dag} & $59.1$ & $78.8$ & $65.7$ & $55.9$ & $47.3$ & $31.0$ & $44.3$ \\
            {\small DeepSeek-V3.0-Chat} \tnote{\dag} & $55.0$ & $80.7$ & $69.7$ & $59.5$ & $49.9$ & $35.0$ & $47.9$ \\
            \midrule
            \multicolumn{8}{c}{\textbf{Mathematical Reasoning LLMs}} \\
            \midrule
            {\small DeepSeek-Math-7B-RL} & $23.5$ & $45.0$ & $29.0$ & $19.8$ & $14.0$ & $9.7$ & $13.7$ \\
            {\small NuminaMath-72B-CoT} & $40.8$ & $63.3$ & $43.5$ & $34.0$ & $27.1$ & $14.2$ & $25.0$ \\
            {\small Qwen2.5-Math-7B-Instruct} & $44.1$ & $68.4$ & $53.0$ & $44.1$ & $38.3$ & $28.1$ & $36.6$ \\
            {\small Qwen2.5-Math-72B-Instruct} & $57.6$ & $74.2$ & $60.4$ & $52.7$ & $45.4$ & $27.9$ & $42.3$ \\
            \midrule
            \multicolumn{8}{c}{\textbf{O1-like Reasoning LLMs}} \\
            \midrule
            {\small Skywork-o1} & $45.4$ & $61.1$ & $47.7$ & $39.3$ & $31.9$ & $21.7$ & $30.4$ \\ 
            {\small QwQ-32B-Preview} & $72.7$ & $89.0$ & $81.8$ & $74.9$ & $65.8$ & $40.1$ & $61.2$ \\
            {\small DeepSeek Distill Qwen-1.5B} & $42.4$ & $83.0$ & $71.7$ & $61.9$ & $48.8$ & $25.6$ & $45.1$ \\
            {\small DeepSeek Distill LLama-8B} & $58.4$ & $88.8$ & $78.4$ & $67.8$ & $56.8$ & $31.9$ & $52.2$ \\
            {\small DeepSeek Distill Qwen-7B} & $65.6$ & $88.5$ & $81.8$ & $73.0$ & $66.4$ & $48.4$ & $63.1$ \\
            {\small OpenAI o1-mini} \tnote{\dag} & $74.1$ \tnote{\ddag} & $89.5$ & $82.4$ & $76.3$ & $67.3$ & $48.3$ & $64.8$ \\
            {\small DeepSeek Distill Qwen-14B} & $69.8$ & $90.5$ & $85.9$ & $79.9$ & $71.2$ & $51.6$ & $68.0$ \\
            {\small DeepSeek Distill Qwen-32B} & $67.7$ & $90.1$ & $86.3$ & $81.2$ & $72.3$ & $54.5$ & $69.7$ \\
            {\small DeepSeek Distill LLama-70B} & $74.8$ & $89.6$ & $86.1$ & $80.8$ & $73.0$ & $53.0$ & $69.7$ \\
            {\small OpenAI o3-mini} \tnote{\dag} & $84.7$ \tnote{\ddag} & $94.3$ & $90.2$ & $85.7$ & $78.8$ & $65.3$ & $76.8$ \\
            {\small DeepSeek R1} \tnote{\dag} & $81.1$ & $91.8$ & $86.9$ & $83.6$ & $79.1$ & $69.5$ & $77.6$ \\
            \bottomrule
        \end{tabular}
        \begin{tablenotes}
            \item[\dag] API-based close-source LLMs.
            \item[\ddag] OpenAI o1-like series model does not provide an optional temperature parameter, so we chose the average accuracy of 20 generations as greedy accuracy.
        \end{tablenotes}
        \end{threeparttable}
    }
\end{table*}

\begin{table*}[t]
    \centering
    \caption{Full performance of models on LiveMathBench hard split. We report results of greedy decoding, Pass@$16$~(G-Pass@$16_{\rightarrow 0}$), G-Pass@$16_{\{0.25, 0.5, 0.75, 1.0\}}$, and mG-Pass@$16$.}
    \label{tab:full_main_performance_hard}
    \resizebox{\textwidth}{!}{
        \begin{threeparttable}[b]
        \begin{tabular}{lccccccc}
            \toprule
            \multirow{2}{*}{\textbf{LLMs}} & \multirow{2}{*}{\textbf{Greedy}} & \multicolumn{6}{c}{\textbf{G-Pass@$16$}~(\Cref{eq:k_at_pass}) / \%} \\
            \cmidrule(lr){3-8}
            & & \textbf{G-Pass@$16_{\rightarrow 0}$} & \textbf{G-Pass@$16_{0.25}$} & \textbf{G-Pass@$16_{0.5}$} & \textbf{G-Pass@$16_{0.75}$} & \textbf{G-Pass@$16_{1.0}$} &  \textbf{mG-Pass@$16$} \\
            \midrule
            \multicolumn{8}{c}{\textbf{General LLMs}} \\
            \midrule
            {\small Llama-3.1-8B-Instruct} & $2.2$ & $35.5$ & $11.5$ & $0.8$ & $0.0$ & $0.0$ & $0.0$ \\
            {\small Qwen2.5-7B-Instruct} & $13.3$ & $23.6$ & $11.4$ & $6.2$ & $3.2$ & $2.2$ & $3.3$ \\
            {\small Llama-3.3-70B-Instruct} & $4.4$ & $23.1$ & $11.0$ & $8.0$ & $4.7$ & $2.3$ & $4.4$ \\
            {\small Llama-3.1-70B-Instruct} & $4.4$ & $25.0$ & $17.5$ & $12.2$ & $7.5$ & $2.7$ & $7.0$ \\
            {\small InternLM3-8B-Instruct} & $11.1$ & $40.0$ & $19.3$ & $10.7$ & $8.2$ & $2.7$ & $7.1$ \\ 
            {\small Qwen2.5-32B-Instruct} & $13.3$ & $30.0$ & $19.1$ & $14.1$ & $10.5$ & $3.5$ & $9.1$ \\
            {\small Qwen2.5-72B-Instruct} & $17.8$ & $29.0$ & $18.1$ & $15.3$ & $11.3$ & $5.4$ & $10.5$ \\
            \midrule
            \multicolumn{8}{c}{\textbf{Mathematical Reasoning LLMs}} \\
            \midrule
            {\small DeepSeek-Math-7B-RL} & $8.9$ & $23.4$ & $6.9$ & $3.9$ & $2.3$ & $0.6$ & $2.2$ \\
            {\small Qwen2.5-Math-7B-Instruct} & $15.6$ & $21.3$ & $13.1$ & $8.2$ & $3.3$ & $2.2$ & $3.8$ \\
            {\small NuminaMath-72B-CoT} & $11.1$ & $28.9$ & $10.3$ & $8.8$ & $7.3$ & $5.9$ & $7.3$ \\
            {\small Qwen2.5-Math-72B-Instruct} & $11.1$ & $32.2$ & $23.2$ & $11.8$ & $7.9$ & $5.9$ & $7.9$ \\
            \midrule
            \multicolumn{8}{c}{\textbf{O1-like Reasoning LLMs}} \\
            \midrule
            {\small QwQ-32B-Preview} & $15.6$ & $54.3$ & $22.1$ & $5.9$ & $4.4$ & $2.4$ & $4.0$ \\
            {\small DeepSeek Distill Qwen-1.5B} & $6.7$ & $33.9$ & $15.8$ & $9.8$ & $4.5$ & $2.2$ & $4.6$ \\
            {\small DeepSeek Distill LLama-8B} & $8.9$ & $56.5$ & $30.7$ & $16.1$ & $5.6$ & $2.4$ & $6.2$ \\
            {\small DeepSeek Distill Qwen-7B} & $17.8$ & $57.4$ & $32.9$ & $13.9$ & $8.8$ & $3.3$ & $8.1$ \\
            {\small OpenAI o1-mini} \tnote{\dag} & $18.4$ \tnote{\ddag} & $68.3$ & $45.6$ & $21.0$ & $10.1$ & $0.5$ & $8.5$ \\
            {\small DeepSeek Distill Qwen-14B} & $15.6$ & $59.8$ & $43.0$ & $26.9$ & $15.9$ & $8.1$ & $15.5$ \\
            {\small DeepSeek Distill Qwen-32B} & $22.2$ & $59.1$ & $43.8$ & $29.9$ & $16.9$ & $3.3$ & $15.1$ \\
            {\small DeepSeek Distill LLama-70B} & $35.6$ & $61.2$ & $50.1$ & $33.1$ & $19.0$ & $5.8$ & $17.3$ \\
            {\small OpenAI o3-mini} \tnote{\dag} & $43.3$ \tnote{\ddag} & $72.3$ & $57.1$ & $47.4$ & $32.5$ & $7.7$ & $28.6$ \\
            {\small DeepSeek R1} \tnote{\dag} & $42.2$ & $71.5$ & $55.9$ & $46.6$ & $33.6$ & $9.8$ & $29.6$ \\
            \bottomrule
        \end{tabular}
        \begin{tablenotes}
            \item[\dag] API-based close-source LLMs.
            \item[\ddag] OpenAI o1-like series model does not provide an optional temperature parameter, so we chose the average accuracy of 20 generations as greedy accuracy.
        \end{tablenotes}
        \end{threeparttable}
    }
\end{table*}

\subsection{Full Performance on MATH500-L5 \& AIME2024-45} \label{app:full_performance_public}

\Cref{tab:full_public_performance}, \Cref{tab:full_aime2025_performance} presents the comprehensive performance results for MATH500-L5, AIME2024-45, and AIME2025.

\begin{table*}[t]
    \centering
    \caption{Full performance of models on MATH500-L5 and AIME2024-45. Results of greedy decoding, Pass@$16$~(G-Pass@$16_{\rightarrow 0}$), G-Pass@$16_{\{0.25, 0.5, 0.75, 1.0\}}$, and mG-Pass@$16$ are reported.}
    \label{tab:full_public_performance}
    \resizebox{\textwidth}{!}{
        \begin{threeparttable}[b]
        \begin{tabular}{lccccccc}
            \toprule
            \multirow{2}{*}{\textbf{LLMs}} & \multirow{2}{*}{\textbf{Greedy}} & \multicolumn{6}{c}{\textbf{G-Pass@$16$}~(\Cref{eq:k_at_pass}) / \%} \\
            \cmidrule(lr){3-8}
            & & \textbf{G-Pass@$16_{\rightarrow 0}$} & \textbf{G-Pass@$16_{0.25}$} & \textbf{G-Pass@$16_{0.5}$} & \textbf{G-Pass@$16_{0.75}$} & \textbf{G-Pass@$16_{1.0}$} & \textbf{mG-Pass@$16$} \\
            \midrule
            \multicolumn{8}{c}{\textbf{MATH500-L5}} \\
            \midrule
            {\small Llama-3.1-8B-Instruct} & $26.1$ & $62.2$ & $29.9$ & $17.8$ & $10.7$ & $3.5$ & $9.7$ \\
            {\small Yi-1.5-34B-Chat} & $26.1$ & $66.4$ & $38.8$ & $25.7$ & $16.7$ & $9.5$ & $16.2$ \\
            {\small Gemma-2-27b-it} & $24.6$ & $50.5$ & $32.6$ & $23.7$ & $18.4$ & $13.0$ & $17.6$ \\ 
            {\small Llama-3.1-70B-Instruct} & $39.6$ & $73.8$ & $53.7$ & $41.8$ & $32.1$ & $16.1$ & $29.3$ \\
            {\small InternLM-3-8B-Instruct} & $51.5$ & $72.3$ & $56.8$ & $49.9$ & $40.3$ & $26.9$ & $38.3$ \\
            {\small Qwen2.5-7B-Instruct} & $56.0$ & $76.1$ & $66.6$ & $54.9$ & $43.3$ & $28.0$ & $41.5$ \\
            {\small Llama-3.3-70B-Instruct} & $54.5$ & $73.1$ & $63.1$ & $55.4$ & $49.5$ & $35.0$ & $47.3$ \\
            {\small Mistral-Large-Instruct-2411} & $55.2$ & $58.4$ & $54.4$ & $52.3$ & $51.2$ & $45.6$ & $50.1$ \\
            {\small Qwen2.5-72B-Instruct} & $63.4$ & $78.9$ & $69.5$ & $62.5$ & $54.4$ & $44.9$ & $53.1$ \\
            {\small Qwen2.5-Max} \tnote{\dag} & $63.4$ & $87.1$ & $73.5$ & $65.8$ & $57.3$ & $38.9$ & $54.5$ \\
            {\small Qwen2.5-32B-Instruct} & $64.2$ & $79.9$ & $71.1$ & $66.6$ & $59.4$ & $41.0$ & $55.6$ \\
            {\small Gemini-1.5-Pro-Latest} \tnote{\dag} & $72.4$ & $92.7$ & $82.7$ & $74.4$ & $64.9$ & $45.3$ & $61.8$ \\
            \hdashline
            {\small DeepSeek-Math-7b-RL} & $15.7$ & $44.0$ & $26.3$ & $15.5$ & $8.7$ & $5.7$ & $9.0$ \\
            {\small NuminaMath-72B-CoT} & $41.0$ & $67.3$ & $51.1$ & $36.8$ & $26.8$ & $16.8$ & $25.6$ \\
            {\small Qwen2.5-Math-72B-Instruct} & $71.6$ & $77.8$ & $71.1$ & $64.9$ & $59.4$ & $46.0$ & $57.4$ \\
            {\small Qwen2.5-Math-7B-Instruct} & $65.7$ & $78.9$ & $71.4$ & $65.0$ & $62.2$ & $57.6$ & $61.5$ \\
            \hdashline
            {\small Skywork-o1} & $61.2$ & $70.9$ & $60.0$ & $56.5$ & $52.2$ & $42.9$ & $50.7$ \\
            {\small DeepSeek Distill Qwen-1.5B} & $53.0$ & $89.5$ & $82.3$ & $72.1$ & $62.1$ & $34.5$ & $57.0$ \\
            {\small QwQ-32B-Preview} & $82.8$ & $95.9$ & $92.5$ & $87.2$ & $78.8$ & $57.4$ & $75.6$ \\
            {\small DeepSeek Distill LLama-8B} & $65.7$ & $92.2$ & $86.6$ & $79.5$ & $70.0$ & $39.5$ & $64.5$ \\
            {\small DeepSeek Distill Qwen-7B} & $78.4$ & $96.8$ & $94.2$ & $87.9$ & $80.5$ & $62.6$ & $77.6$ \\
            {\small DeepSeek Distill Qwen-14B} & $76.1$ & $97.1$ & $93.4$ & $91.1$ & $85.9$ & $67.7$ & $82.6$ \\
            {\small DeepSeek Distill LLama-70B} & $87.3$ & $96.5$ & $93.5$ & $89.6$ & $85.5$ & $66.8$ & $81.9$ \\
            {\small DeepSeek Distill Qwen-32B} & $83.6$ & $96.1$ & $93.6$ & $89.9$ & $83.8$ & $70.4$ & $81.9$ \\
            \midrule
            \multicolumn{8}{c}{\textbf{AIME2024-45}} \\
            \midrule
            {\small Yi-1.5-34B-Chat} & $2.2$ & $20.5$ & $5.9$ & $0.5$ & $0.0$ & $0.0$ & $0.0$ \\
            {\small Llama-3.1-8B-Instruct} & $4.4$ & $28.1$ & $4.9$ & $2.2$ & $1.6$ & $0.0$ & $1.2$ \\
            {\small Gemma-2-27b-it} & $6.7$ & $21.0$ & $8.3$ & $5.2$ & $1.8$ & $0.0$ & $1.8$ \\ 
            {\small InternLM-3-8B-Instruct} & $11.1$ & $20.5$ & $13.3$ & $7.2$ & $4.3$ & $1.0$ & $3.7$ \\
            {\small Qwen2.5-32B-Instruct} & $11.1$ & $32.0$ & $14.9$ & $7.1$ & $3.4$ & $2.2$ & $3.7$ \\
            {\small Mistral-Large-Instruct-2411} & $13.3$ & $15.4$ & $11.1$ & $10.4$ & $6.8$ & $2.4$ & $6.1$ \\
            {\small Qwen2.5-7B-Instruct} & $11.1$ & $26.3$ & $11.2$ & $8.9$ & $8.1$ & $4.7$ & $7.5$ \\
            {\small Llama-3.1-70B-Instruct} & $15.6$ & $41.2$ & $23.5$ & $15.0$ & $8.1$ & $3.0$ & $8.0$ \\
            {\small Gemini-1.5-Pro-Latest} \tnote{\dag} & $13.3$ & $45.7$ & $26.5$ & $16.8$ & $8.7$ & $2.5$ & $8.1$ \\
            {\small Qwen2.5-72B-Instruct} & $13.3$ & $33.7$ & $16.3$ & $13.7$ & $12.9$ & $7.5$ & $11.7$ \\
            {\small Llama-3.3-70B-Instruct} & $22.2$ & $37.1$ & $28.7$ & $25.3$ & $18.2$ & $6.9$ & $16.4$ \\
            {\small Qwen2.5-Max} \tnote{\dag} & $22.2$ & $44.4$ & $25.2$ & $15.5$ & $9.9$ & $5.3$ & $9.8$ \\
            \hdashline
            {\small DeepSeek-Math-7b-RL} & $2.2$ & $16.3$ & $4.4$ & $1.5$ & $0.1$ & $0.0$ & $0.1$ \\
            {\small NuminaMath-72B-CoT} & $2.2$ & $21.3$ & $4.8$ & $2.9$ & $2.2$ & $0.1$ & $1.6$ \\
            {\small Qwen2.5-Math-7B-Instruct} & $11.1$ & $20.8$ & $8.5$ & $4.6$ & $2.6$ & $2.2$ & $2.8$ \\
            {\small Qwen2.5-Math-72B-Instruct} & $20.0$ & $35.2$ & $24.8$ & $18.7$ & $16.2$ & $6.7$ & $14.1$ \\
            \hdashline
            {\small Skywork-o1} & $11.1$ & $22.1$ & $13.6$ & $11.2$ & $10.3$ & $1.5$ & $8.2$ \\
            {\small DeepSeek Distill Qwen-1.5B} & $17.8$ & $68.7$ & $41.6$ & $23.9$ & $14.9$ & $2.8$ & $13.5$ \\
            {\small QwQ-32B-Preview} & $44.4$ & $74.3$ & $59.3$ & $41.0$ & $28.6$ & $8.1$ & $24.7$ \\
            {\small DeepSeek Distill LLama-8B} & $44.4$ & $82.1$ & $72.6$ & $53.9$ & $30.4$ & $9.0$ & $28.0$ \\
            {\small DeepSeek Distill Qwen-7B} & $44.4$ & $79.6$ & $73.9$ & $56.3$ & $35.4$ & $17.5$ & $33.8$ \\
            {\small OpenAI o1-mini} \tnote{\dag} & $60.3$ \tnote{\ddag} & $86.7$ & $80.0$ & $62.2$ & $53.3$ & $15.6$ & $43.1$ \\
            {\small DeepSeek Distill Qwen-14B} & $62.2$ & $86.5$ & $79.3$ & $75.8$ & $62.9$ & $26.5$ & $56.0$ \\
            {\small DeepSeek Distill LLama-70B} & $62.2$ & $84.4$ & $76.9$ & $72.9$ & $63.4$ & $32.2$ & $57.6$ \\
            {\small DeepSeek Distill Qwen-32B} & $62.2$ & $86.3$ & $79.7$ & $77.0$ & $66.5$ & $31.3$ & $59.3$ \\
            \bottomrule
        \end{tabular} 
        \begin{tablenotes}
            \item[\dag] API-based LLMs.
            \item[\ddag] OpenAI o1 series model does not provide an optional temperature parameter, so we chose the average accuracy of 20 generations as greedy accuracy.
        \end{tablenotes}
        \end{threeparttable}
    }
\end{table*}

\begin{table*}[t]
    \centering
    \caption{Full performance of models on AIME2025. We report results of greedy decoding, Pass@$16$~(G-Pass@$16_{\rightarrow 0}$), G-Pass@$16_{\{0.25, 0.5, 0.75, 1.0\}}$, and mG-Pass@$16$.}
    \label{tab:full_aime2025_performance}
    \resizebox{\textwidth}{!}{
        \begin{threeparttable}[b]
        \begin{tabular}{lccccccc}
            \toprule
            \multirow{2}{*}{\textbf{LLMs}} & \multirow{2}{*}{\textbf{Greedy}} & \multicolumn{6}{c}{\textbf{G-Pass@$16$}~(\Cref{eq:k_at_pass}) / \%} \\
            \cmidrule(lr){3-8}
            & & \textbf{G-Pass@$16_{\rightarrow 0}$} & \textbf{G-Pass@$16_{0.25}$} & \textbf{G-Pass@$16_{0.5}$} & \textbf{G-Pass@$16_{0.75}$} & \textbf{G-Pass@$16_{1.0}$} &  \textbf{mG-Pass@$16$} \\
            \midrule
            \multicolumn{8}{c}{\textbf{General LLMs}} \\
            \midrule
            {\small Llama-3.1-8B-Instruct} & $0.0$ & $8.9$ & $0.0$ & $0.0$ & $0.0$ & $0.0$ & $0.0$ \\
            {\small Gemma-2-27b-it} & $0.0$ & $9.5$ & $0.0$ & $0.0$ & $0.0$ & $0.0$ & $0.0$ \\
            {\small Yi-1.5-34B-Chat} & $0.0$ & $14.8$ & $4.8$ & $0.1$ & $0.0$ & $0.0$ & $0.0$ \\
            {\small GPT-4o-2024-11-20} \tnote{\dag} & $0.0$ & $25.5$ & $7.4$ & $0.1$ & $0.0$ & $0.0$ & $0.0$ \\
            {\small Llama-3.1-70B-Instruct} & $6.7$ & $21.3$ & $8.3$ & $4.6$ & $0.2$ & $0.0$ & $0.7$ \\
            {\small InternLM3-8B-Instruct} & $13.3$ & $30.3$ & $16.2$ & $6.7$ & $0.1$ & $0.0$ & $0.8$ \\ 
            {\small Qwen2.5-32B-Instruct} & $20.0$ & $33.3$ & $28.1$ & $11.5$ & $0.2$ & $0.0$ & $1.4$ \\
            {\small Claude-3.5-Sonnet} \tnote{\dag} & $13.3$ & $34.4$ & $16.9$ & $6.4$ & $1.2$ & $0.0$ & $1.7$ \\
            {\small Qwen2.5-7B-Instruct} & $6.7$ & $25.2$ & $13.5$ & $9.7$ & $6.2$ & $0.2$ & $4.7$ \\
            {\small Qwen2.5-72B-Instruct} & $20.0$ & $33.2$ & $23.4$ & $12.2$ & $5.8$ & $0.1$ & $4.9$ \\
            {\small Llama-3.3-70B-Instruct} & $6.7$ & $13.6$ & $6.7$ & $6.7$ & $6.6$ & $0.5$ & $5.0$ \\
            {\small Gemini-1.5-Pro-Latest} \tnote{\dag} & $20.0$ & $40.9$ & $25.8$ & $10.8$ & $6.7$ & $4.4$ & $6.8$ \\
            {\small Qwen2.5-Max} \tnote{\dag} & $13.3$ & $39.9$ & $24.0$ & $11.9$ & $6.8$ & $2.9$ & $6.8$ \\
            {\small Mistral-Large-Instruct-2411} & $13.3$ & $19.7$ & $14.4$ & $10.8$ & $6.8$ & $6.7$ & $7.2$ \\
            {\small Gemini-2.0-Flash-Exp} \tnote{\dag} & $26.7$ & $44.7$ & $30.3$ & $26.5$ & $21.5$ & $14.0$ & $21.2$ \\
            \midrule
            \multicolumn{8}{c}{\textbf{Mathematical Reasoning LLMs}} \\
            \midrule
            {\small DeepSeek-Math-7B-RL} & $0.0$ & $0.0$ & $0.0$ & $0.0$ & $0.0$ & $0.0$ & $0.0$ \\
            {\small NuminaMath-72B-CoT} & $0.0$ & $21.0$ & $10.3$ & $6.7$ & $6.7$ & $4.4$ & $6.4$ \\
            {\small Qwen2.5-Math-7B-Instruct} & $20.0$ & $36.8$ & $16.2$ & $8.7$ & $6.7$ & $6.7$ & $6.8$ \\
            {\small Qwen2.5-Math-72B-Instruct} & $13.3$ & $30.1$ & $15.7$ & $13.3$ & $13.3$ & $13.3$ & $13.3$ \\
            \midrule
            \multicolumn{8}{c}{\textbf{O1-like Reasoning LLMs}} \\
            \midrule
            {\small Skywork-o1} & $13.3$ & $31.2$ & $21.5$ & $15.3$ & $13.3$ & $7.2$ & $11.8$ \\ 
            {\small DeepSeek Distill Qwen-1.5B} & $26.7$ & $54.6$ & $38.1$ & $31.9$ & $23.8$ & $1.4$ & $18.6$ \\
            {\small DeepSeek Distill LLama-8B} & $40.0$ & $62.2$ & $55.3$ & $40.4$ & $21.2$ & $7.9$ & $21.0$ \\
            {\small QwQ-32B-Preview} & $26.7$ & $60.5$ & $43.5$ & $34.5$ & $32.4$ & $15.6$ & $28.1$ \\
            {\small OpenAI o1-mini} \tnote{\dag} & $46.7$ \tnote{\ddag} & $62.0$ & $41.0$ & $39.9$ & $32.5$ & $14.0$ & $28.4$ \\
            {\small DeepSeek Distill Qwen-7B} & $46.7$ & $62.1$ & $53.2$ & $46.6$ & $38.3$ & $22.7$ & $36.1$ \\
            {\small DeepSeek Distill Qwen-14B} & $46.7$ & $67.7$ & $60.1$ & $58.8$ & $41.1$ & $25.2$ & $40.8$ \\
            {\small DeepSeek Distill LLama-70B} & $46.7$ & $75.1$ & $54.9$ & $52.5$ & $38.6$ & $26.8$ & $37.4$ \\
            {\small DeepSeek R1} \tnote{\dag} & $66.7$ & $68.9$ & $63.7$ & $52.6$ & $46.8$ & $24.3$ & $42.5$ \\
            {\small OpenAI o3-mini} \tnote{\dag} & $53.3$ \tnote{\ddag} & $80.0$ & $77.7$ & $59.0$ & $46.5$ & $29.4$ & $43.6$ \\
            {\small DeepSeek Distill Qwen-32B} & $46.7$ & $72.6$ & $63.6$ & $59.7$ & $50.2$ & $29.5$ & $47.3$ \\
            \bottomrule
        \end{tabular}
        \begin{tablenotes}
            \item[\dag] API-based close-source LLMs.
            \item[\ddag] OpenAI o1-like series model does not provide an optional temperature parameter, so we chose the average accuracy of 20 generations as greedy accuracy.
        \end{tablenotes}
        \end{threeparttable}
    }
\end{table*}

\subsection{Full Performance \textit{w.r.t.} Different temperatures} \label{app:full_performance_temp}
\begin{figure*}[ht]
    \centering
    \includegraphics[width=\linewidth]{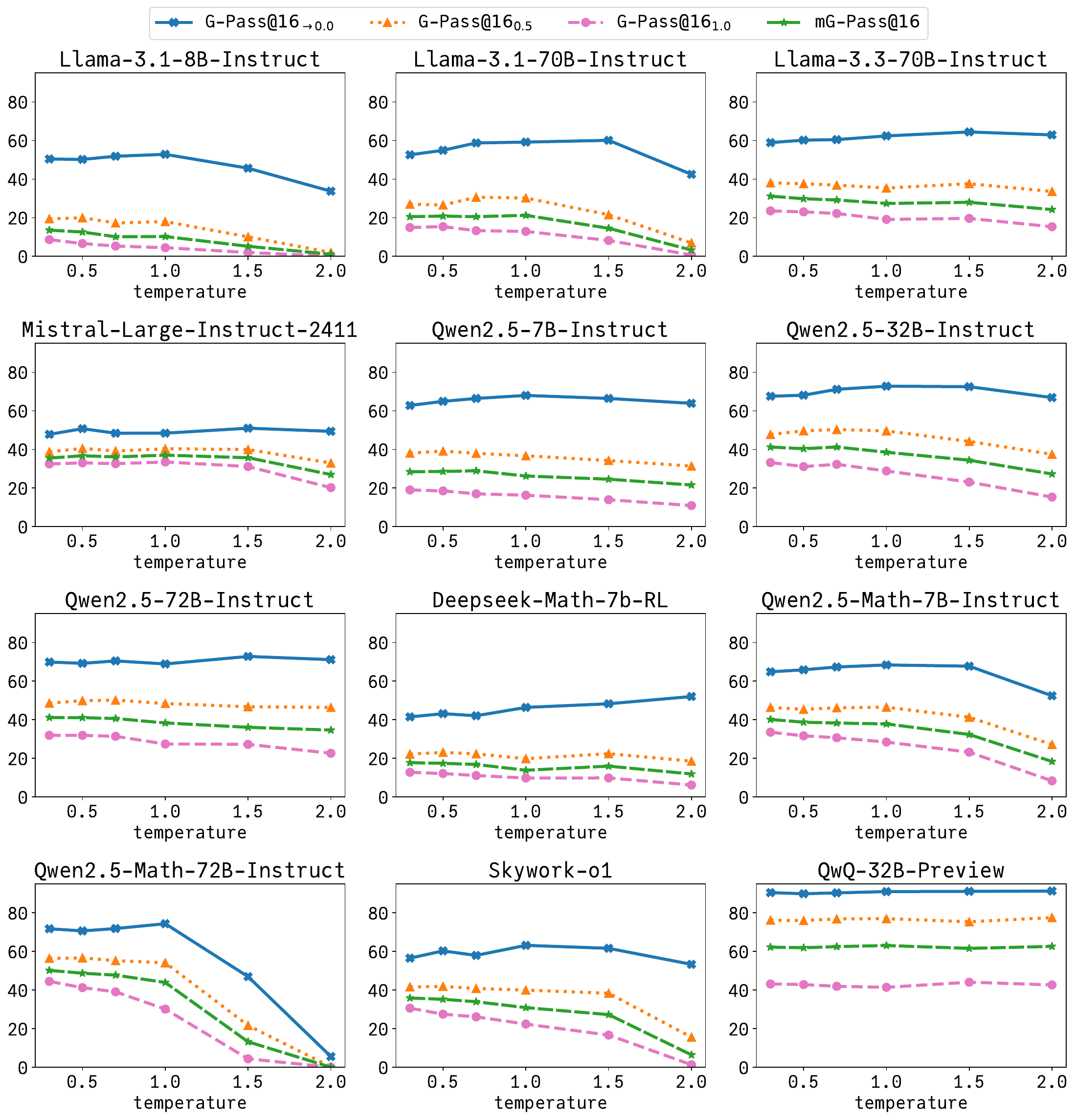}
    \caption{\gpkt performance of LLMs \textit{w.r.t.} different temperatures.}
    \label{fig:g-pass-at-k-wrt-t-full}
\end{figure*}

\Cref{fig:g-pass-at-k-wrt-t-full} demonstrates the performance of models across different sampling temperatures. 
The values for the other sampling parameters, namely $\text{top-p}$~(0.8), $\text{top-k}$~(50), and $\text{repetition-penalty}$~(1.0), remain constant. 
It can be observed that most models exhibit relatively stable performance when the temperature values are set to $\{0.3, 0.5, 0.7, 1.0\}$.
This indicates that the \gpkt metric is robust for commonly used temperature parameters. 
However, it is worth noting that certain models, such as Qwen2.5-Math-7B-Instruct and Qwen2.5-Math-72B-Instruct, display an unusual sensitivity to larger temperature coefficients. 
This discrepancy may reflect the impact of different training strategies on the model distribution. 
Intriguingly, we discover that the model QwQ-32B-Preview shows exceptional stability when it comes to sampling temperatures. 
We speculate that models similar to o1, which rely on long COT and reflection, are capable of consistently identifying the correct answers through continuous self-examination and backtracking. 
This characteristic contributes to the high reasoning stability observed in these models.

\subsection{Full Performance \textit{w.r.t.} Different top-p} \label{app:full_performance_topp}
\begin{figure*}[ht]
    \centering
    \includegraphics[width=\linewidth]{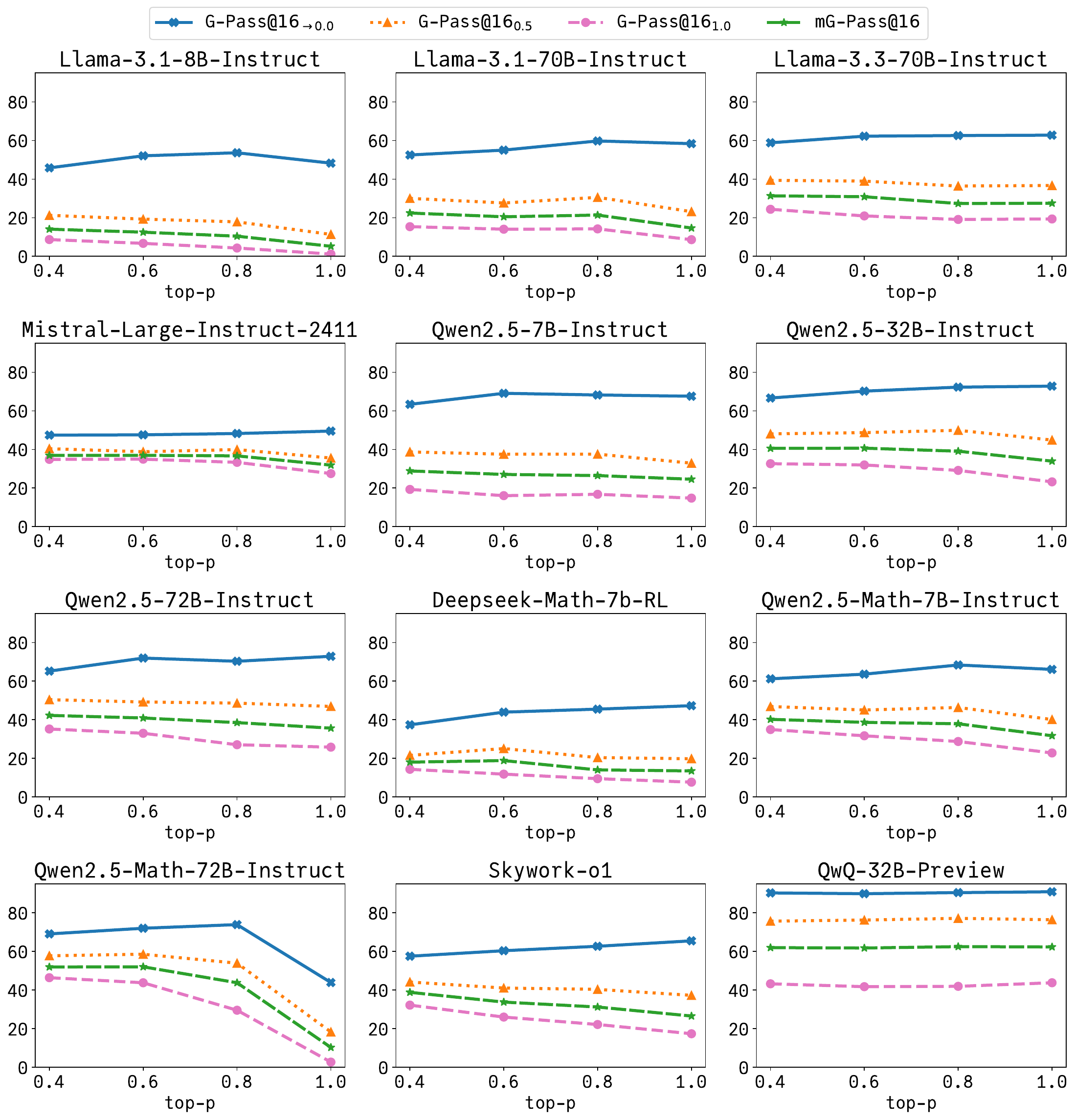}
    \caption{\gpkt performance of LLMs \textit{w.r.t.} different $\text{top-p}$.}
    \label{fig:g-pass-at-k-wrt-p-full}
\end{figure*}

\Cref{fig:g-pass-at-k-wrt-p-full} illustrates the performance of models across various sampling $\text{top-p}$ parameters. 
The values for the other sampling parameters, namely, $\text{temperature}$~(1.0), $\text{top-k}$~(50), and $\text{repetition-penalty}$~(1.0), remain constant. 
Similar phenomena are also observed in the experiments conducted with varying temperatures. 
Most models exhibit stable performance within the range of commonly used parameters, demonstrating the effectiveness of \gpkt. 
Furthermore, QwQ-32B-Preview also exhibits stability \textit{w.r.t.} different $\text{top-p}$ values.

\subsection{Full Performance \textit{w.r.t.} Different top-k} \label{app:full_performance_topk}
\begin{figure*}[ht]
    \centering
    \includegraphics[width=\linewidth]{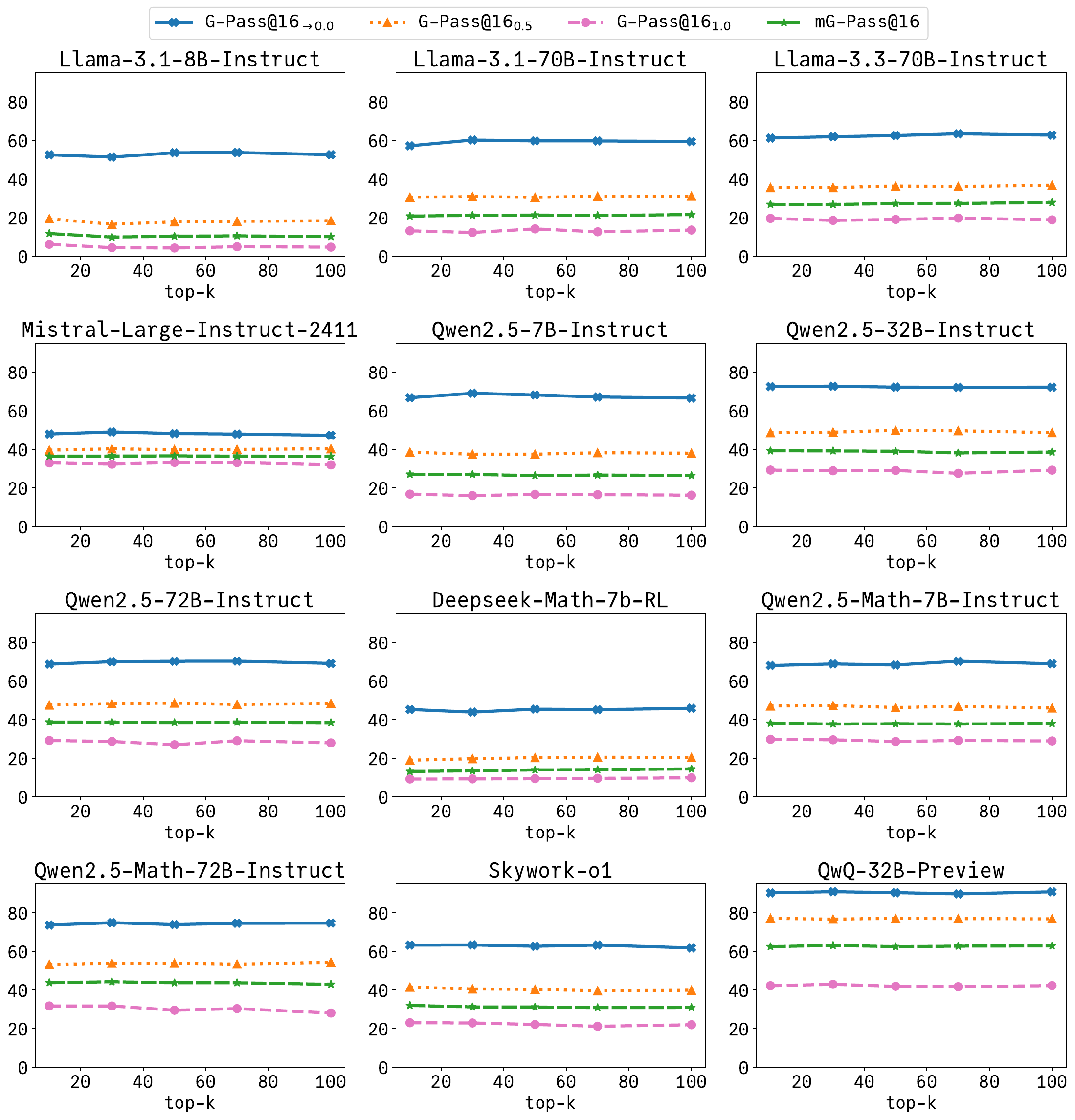}
    \caption{\gpkt performance of LLMs \textit{w.r.t.} different $\text{top-k}$.}
    \label{fig:g-pass-at-k-wrt-k-full}
\end{figure*}

\Cref{fig:g-pass-at-k-wrt-k-full} illustrates the performance of models across various sampling $\text{top-k}$ parameters. 
The values for the other sampling parameters, specifically, temperature~(1.0), $\text{top-p}$~(0.8), and $\text{repetition-penalty}$~(1.0), remain constant. 
Our analysis indicates that all models exhibit stable performance, suggesting that the $\text{top-k}$ parameter has a minimal effect on sampling compared to temperature and $\text{top-p}$.

\clearpage

\begin{figure*}[!ht] 
    \begin{AIbox}{Question 1}
        In $\triangle ABC$, given $\cos C = \frac{\sin A + \cos A}{2} = \frac{\sin B + \cos B}{2}$, find the value of $\cos C$.
        \par
    \end{AIbox}
\end{figure*}

\begin{figure*}[!ht] 
    \begin{AIbox}{QwQ-32B-Preview Response for Question 1}

        Let's tackle this problem step by step. So, we have a triangle ABC, and we're given that $\cos C$ equals both $(\sin A + \cos A)/2$ and $(\sin B + \cos B)/2$. Our goal is to find the value of $\cos C$.

        First, let's write down what we know:

        \[
        \cos C = \frac{\sin A + \cos A}{2} = \frac{\sin B + \cos B}{2}
        \]
        
        

        .........
        

        Therefore, $\cos C = (\sqrt{7})/4$. \par

        \colorbox{Turquoise}{Wait, but earlier steps led to cos C = -cos 2A, and now cos $C = (\sqrt{7})/4$.} \par

        Let me verify if this makes sense. \par

        We have $\cos C = (\sqrt{7})/4$, and from $\cos C = -\cos 2A$. \par

        So, $-\cos 2A = (\sqrt{7})/4 \rightarrow cos 2A = -\sqrt{7}/4$. \par
 
        \colorbox{Turquoise}{But $\cos 2A = 2 \cos^2A - 1$.} \par

        Substitute $\cos A = ( \sqrt{7} - 1 ) / 4$: \par
 
        \[
        \begin{aligned}
            2 \left( \frac{\sqrt{7} - 1}{4} \right)^2 - 1 = 2 \left( \frac{7 - 2\sqrt{7} + 1}{16} \right) - 1 = 2 \left( \frac{8 - 2\sqrt{7}}{16} \right) - 1 = \frac{8 - 2\sqrt{7}}{8} - 1 \\= \frac{8 - 2\sqrt{7} - 8}{8} = -\frac{\sqrt{7}}{4}
        \end{aligned}
        \]

        This matches $\cos C = -\cos 2A = - ( -\sqrt{7}/4 ) = \sqrt{7}/4$. \par

        Therefore, $\cos C = \sqrt{7}/4$. \par

        \colorbox{Turquoise}{Wait, but I thought $\cos C = (\sqrt{7})/4$, and $\cos 2A = -\sqrt{7}/4$, which satisfies $\cos C = -\cos 2A$.} \par

        Therefore, the value of $\cos C$ is $\sqrt{7}/4$. \par

        So, the answer should be \boxed{\dfrac{\sqrt{7}}{4}}. \par

        \textbf{Final Answer} \par

        \boxed{\dfrac{\sqrt{7}}{4}} \par
    \end{AIbox}
\end{figure*}

\begin{figure*}[!ht] 
    \begin{AIbox}{Qwen2.5-32B-Instruct-Preview Response Question 1}
        Given the equation in $\triangle ABC$: 
        \[ \cos C = \frac{\sin A + \cos A}{2} = \frac{\sin B + \cos B}{2} \]
        
        We will denote the common value of these expressions by $x$:
        \[ \cos C = x \]
        \[ \frac{\sin A + \cos A}{2} = x \]
        \[ \frac{\sin B + \cos B}{2} = x \]
        
        From these equations, we derive:
        \[ \sin A + \cos A = 2x \]
        \[ \sin B + \cos B = 2x \]
        
        1. \textbf{Square both sides of each equation:}
        
        \[
        (\sin A + \cos A)^2 = (2x)^2
        \]
        \[
        (\sin B + \cos B)^2 = (2x)^2
        \]
        
        2. \textbf{Expand the squares:}
        
        \[
        \sin^2 A + \cos^2 A + 2 \sin A \cos A = 4x^2
        \]
        \[
        \sin^2 B + \cos^2 B + 2 \sin B \cos B = 4x^2
        \]
        
        3. \textbf{Use the Pythagorean identity $\sin^2 \theta + \cos^2 \theta = 1$:}
        
        \[
        1 + 2 \sin A \cos A = 4x^2
        \]
        \[
        1 + 2 \sin B \cos B = 4x^2
        \]
        
        4. \textbf{Subtract 1 from both sides:}
        
        \[
        2 \sin A \cos A = 4x^2 - 1
        \]
        \[
        2 \sin B \cos B = 4x^2 - 1
        \]
        
        ..............
        
        16. \textbf{Since $\sin A = \cos A$ when $A = B$:}
        
        \[
        \sin A = \cos A \implies \sin A = \frac{\sqrt{2}}{2}
        \]
        \[
        2 \sin^2 A = 1
        \]
        
        17. \textbf{Substitute back to find $x$:}
        
        \[
        x = -1 + 1 = 0
        \]
        
        Thus, the value of $\cos C$ is:
        \[
        \boxed{\frac{\sqrt{2}}{2}}
        \]
        
    \end{AIbox}
\end{figure*}

\clearpage

\begin{figure*}[t]
    \centering
    \includegraphics[scale=.5]{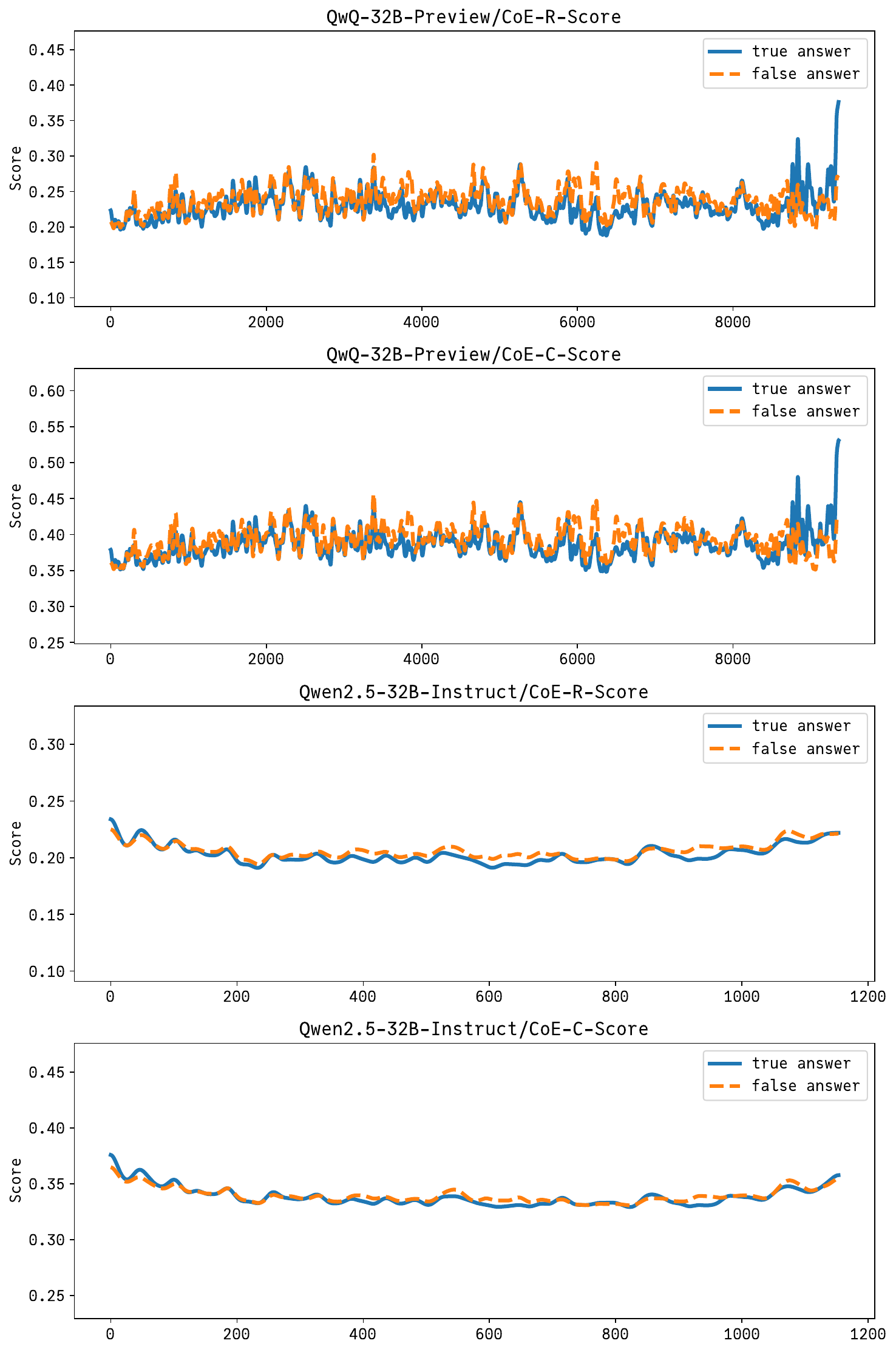}
    \caption{CoE-Score of QwQ and Qwen on Question 1.}
    \label{fig:coe_score1}
\end{figure*}

\newpage

\begin{figure*}[!ht] 
    \begin{AIbox}{Question 2}
        The sequence $\{a_n\}$ satisfies $a_1=1$, and for any positive integer $n$, we have $a_{n+1}=10^{n}{a_n^2}$. What is the general term formula for $\{a_n\}$?
        \par
    \end{AIbox}
\end{figure*}

\begin{figure*}[!ht] 
    \begin{AIbox}{QwQ-32B-Preview Response for Question 2}

        Let's tackle this problem step by step. We have a sequence defined by a recursive formula, and we need to find a general term for it. The sequence is given by \( a_1 = 1 \), and for any positive integer \( n \), \( a_{n+1} = 10^n a_n^2 \). Our goal is to find a formula for \( a_n \) in terms of \( n \).

        First, I'll try to compute the first few terms of the sequence to see if there's a pattern.
        
        Given:
        \[ a_1 = 1 \]
        
        For \( n = 1 \):
        \[ a_2 = 10^1 \cdot a_1^2 = 10 \cdot (1)^2 = 10 \]
        
        For \( n = 2 \):
        \[ a_3 = 10^2 \cdot a_2^2 = 100 \cdot (10)^2 = 100 \cdot 100 = 10,000 \]
        
        For \( n = 3 \):
        \[ a_4 = 10^3 \cdot a_3^2 = 1,000 \cdot (10,000)^2 = 1,000 \cdot 100,000,000 = 10^{12} \]

        .........
        
        \colorbox{Turquoise}{Wait a minute, let's check this with the earlier terms to verify.}
        
        For \( n = 1 \):
        
        \[ a_1 = 10^{2^1 - 1 - 1} = 10^{2 - 1 - 1} = 10^{0} = 1 \]
        
        Correct.
        
        For \( n = 2 \):
        
        \[ a_2 = 10^{2^2 - 2 - 1} = 10^{4 - 2 - 1} = 10^{1} = 10 \]
        
        Correct.
        
        For \( n = 3 \):
        
        \[ a_3 = 10^{2^3 - 3 - 1} = 10^{8 - 3 - 1} = 10^{4} = 10,000 \]
        
        Correct.
        
        For \( n = 4 \):
        
        \[ a_4 = 10^{2^4 - 4 - 1} = 10^{16 - 4 - 1} = 10^{11} = 100,000,000,000 \]
        
        \colorbox{Turquoise}{Wait, but earlier I calculated \( a_4 = 10^{12} \). Hmm, there's a discrepancy here.}
        
        \colorbox{Turquoise}{Wait, let me double-check my earlier calculation for \( a_4 \):}
        
        Given \( a_3 = 10^4 \), then:
        
        \[ a_4 = 10^3 \cdot (10^4)^2 = 10^3 \cdot 10^8 = 10^{11} \]
        
        \colorbox{Turquoise}{Ah, I made a mistake earlier.} I thought \( a_4 = 10^{12} \), but actually, it's \( 10^{11} \). So, my general formula \( a_n = 10^{2^n - n - 1} \) gives \( a_4 = 10^{11} \), which matches the correct calculation. Great!
        
        Therefore, the general term formula for the sequence \( \{a_n\} \) is:
        
        \[ a_n = 10^{2^{n} - n - 1} \]
        
        \textbf{Final Answer}
        
        \[ \boxed{a_n = 10^{2^{n} - n - 1}} \]
    \end{AIbox}
\end{figure*}

\begin{figure*}[!ht] 
    \begin{AIbox}{Qwen2.5-32B-Instruct-Preview Response Question 2}
        To find the general term formula for the sequence \(\{a_n\}\) defined by \(a_1 = 1\) and \(a_{n+1} = 10^n a_n^2\) for any positive integer \(n\), we will proceed step by step.

        1. \textbf{Initial Condition:} $ a_1 = 1 $
        
        2. \textbf{First Iteration:} $ a_2 = 10^1 a_1^2 = 10 \cdot 1^2 = 10 $
        
        3. \textbf{Second Iteration:} $ a_3 = 10^2 a_2^2 = 10^2 \cdot 10^2 = 10^4 $
        
        4. \textbf{Third Iteration:} $ a_4 = 10^3 a_3^2 = 10^3 \cdot (10^4)^2 = 10^3 \cdot 10^8 = 10^{11} $
        
        From these calculations, we observe a pattern in the exponents of \(10\). Let's hypothesize the general form of \(a_n\) and then prove it by induction.
        
        5. \textbf{Hypothesis:}
           Suppose \(a_n = 10^{b_n}\), where \(b_n\) is some function of \(n\).
        
        6. \textbf{Base Case:}
           For \(n = 1\),
           \[
           a_1 = 1 = 10^0 \implies b_1 = 0
           \]
        
        7. \textbf{Inductive Step:}
           Assume \(a_k = 10^{b_k}\) for some \(k \geq 1\). We need to show that \(a_{k+1} = 10^{b_{k+1}}\).
        
           By the recurrence relation,
           \[
           a_{k+1} = 10^k a_k^2 = 10^k (10^{b_k})^2 = 10^k \cdot 10^{2b_k} = 10^{k + 2b_k}
           \]
           Thus, if \(a_k = 10^{b_k}\), then \(a_{k+1} = 10^{b_{k+1}}\) with \(b_{k+1} = k + 2b_k\).
        
        8. \textbf{Solving the Recurrence for \(b_n\):}
           We have the recurrence relation:
           \[
           b_{k+1} = k + 2b_k
           \]
           with the initial condition \(b_1 = 0\).
        
           To solve this, we will find a pattern by calculating the first few terms:
           \[
           b_1 = 0
           \]
           \[
           b_2 = 1 + 2b_1 = 1 + 2 \cdot 0 = 1
           \]
           \[
           b_3 = 2 + 2b_2 = 2 + 2 \cdot 1 = 4
           \]
           \[
           b_4 = 3 + 2b_3 = 3 + 2 \cdot 4 = 11
           \]
        
           Observing the pattern, we hypothesize:
           \[
           b_n = \frac{2^n - n - 2}{3}
           \]
        
        9. \textbf{Verification:}
           We verify by induction:
           - Base case \(n = 1\):
             \[
             b_1 = \frac{2^1 - 1 - 2}{3} = \frac{2 - 1 - 2}{3} = \frac{-1}{3} = 0
             \]
           - Inductive step:
             Assume \(b_k = \frac{2^k - k - 2}{3}\). Then,
             \[
             b_{k+1} = k + 2b_k = k + 2 \left( \frac{2^k - k - 2}{3} \right) = \frac{3k + 2^{k+1} - 2k - 4}{3} = \frac{2^{k+1} + k - 4}{3}
             \]
             which matches our hypothesis.
        
        Therefore, the general term of the sequence is:
        \[
        a_n = 10^{b_n} = 10^{\frac{2^n - n - 2}{3}}
        \]
        
        The final answer is:
        \[
        \boxed{10^{\frac{2^n - n - 2}{3}}}
        \]
        
    \end{AIbox}
\end{figure*}

\clearpage

\begin{figure*}[t]
    \centering
    \includegraphics[scale=.5]{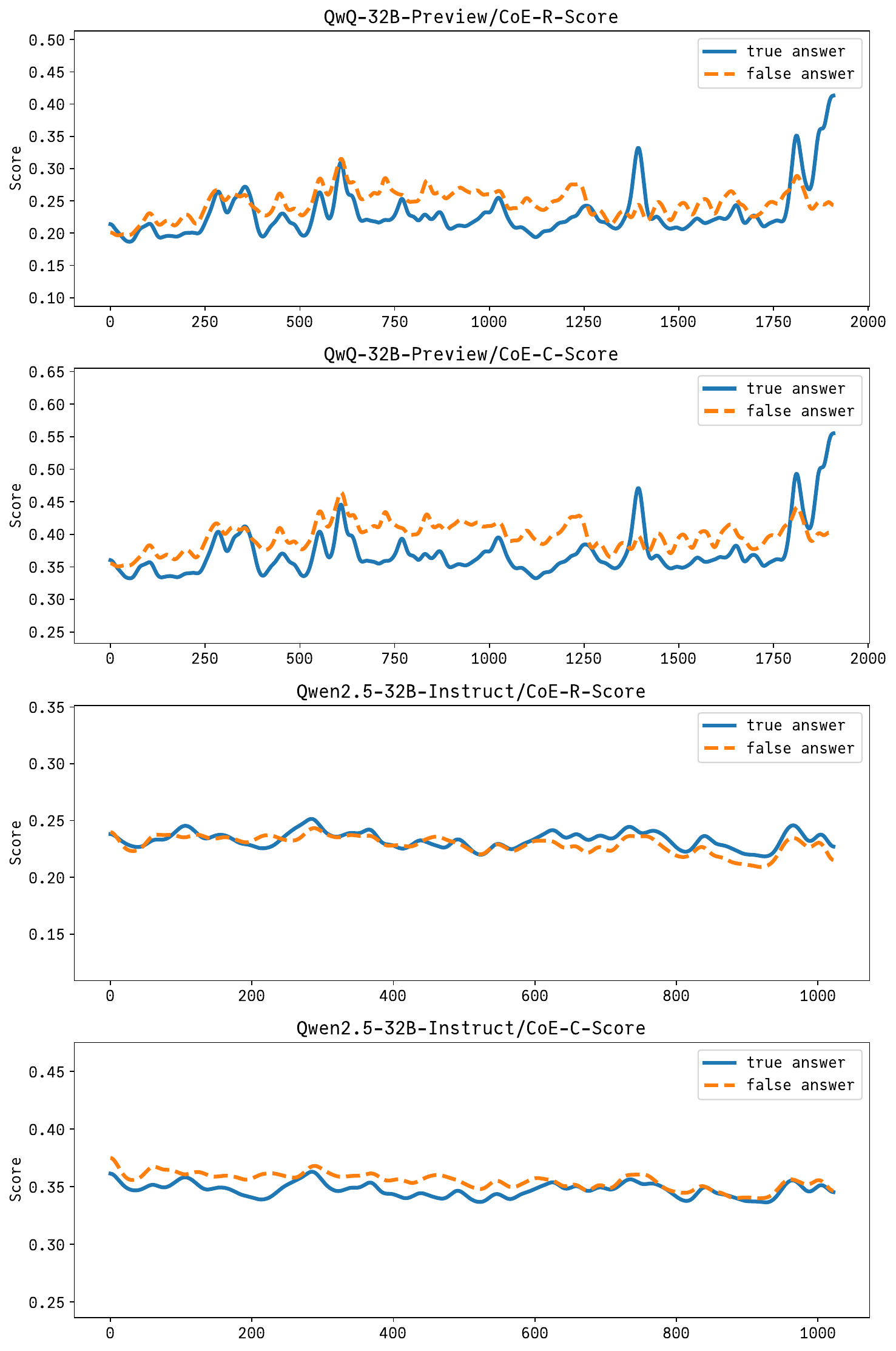}
    \caption{CoE-Score of QwQ and Qwen on Question 2.}
    \label{fig:coe_score2}
\end{figure*}

\newpage

\begin{figure*}[!ht] 
    \begin{AIbox}{Question 3}
        If the real part of the complex number $z$ is greater than 0, and $\overline{z}(z+1)=\frac{20}{3+i}$, then $z=$?
        \par
    \end{AIbox}
\end{figure*}

\begin{figure*}[!ht] 
    \begin{AIbox}{QwQ-32B-Preview Response for Question 3}

        Let's tackle this problem step by step. So, we have a complex number \( z \) with a positive real part, and it satisfies the equation \( \overline{z}(z + 1) = \frac{20}{3 + i} \). We need to find what \( z \) is.

        First, let's recall that for any complex number \( z = a + bi \), where \( a \) and \( b \) are real numbers, the complex conjugate \( \overline{z} \) is \( a - bi \).
        
        So, if \( z = a + bi \), then \( \overline{z} = a - bi \).
        
        Given that, the equation becomes:
        
        \( (a - bi)( (a + bi) + 1 ) = \frac{20}{3 + i} \)
        
        Simplify the left side:
        
        First, compute \( z + 1 = (a + bi) + 1 = (a + 1) + bi \).
        
        Then, \( \overline{z}(z + 1) = (a - bi)((a + 1) + bi) \).
        
        Let's multiply these two complex numbers:
        
        \( (a - bi)( (a + 1) + bi ) = a(a + 1) + a(bi) - bi(a + 1) - bi(bi) \)
        
        \( = a(a + 1) + ab i - ba i - bi + b^2 i^2 \)
        
        \colorbox{Turquoise}{Wait, that seems messy. Maybe there's a better way to expand this.}

        Actually, let's use the distributive property more carefully:

        \( (a - bi)( (a + 1) + bi ) = a(a + 1) + a(bi) - bi(a + 1) - bi(bi) \)
        
        Now, compute each term:
        
        \( a(a + 1) = a^2 + a \)
        
        \( a(bi) = ab i \)
        
        \( -bi(a + 1) = -ba i - b i \)
        
        \( -bi(bi) = -b^2 i^2 = -b^2 (-1) = b^2 \)
        
        Now, combine all these:
        
        \( a^2 + a + ab i - ba i - b i + b^2 \)
        
        Notice that \( ab i - ba i = 0 \), since they cancel each other out.
        
        So, we're left with:
        
        \( a^2 + a + b^2 - b i \)
        
        \colorbox{Turquoise}{Wait, but that doesn't seem right} because the imaginary parts should cancel out completely since the right side is a real number (as we'll see when we simplify it).
        
        \colorbox{Turquoise}{Wait, actually, no. Let's check the right side:}
        
        \( \frac{20}{3 + i} \)
        
        ............................
        
        \colorbox{Turquoise}{However, the problem states that the real part of \( z \) is greater than 0. So, \( a > 0 \).}
        
        Therefore, \( a = 1 \) (since \( a = -2 \) is less than 0 and doesn't satisfy the condition).
        
        So, \( z = a + bi = 1 + 2i \).
        
        Let me verify this solution by plugging back into the original equation.
        
        Compute \( \overline{z}(z + 1) \):
        
        \( \overline{z} = 1 - 2i \)
        
        \( z + 1 = (1 + 2i) + 1 = 2 + 2i \)
        
        Now, \( \overline{z}(z + 1) = (1 - 2i)(2 + 2i) \)
        
        Compute this product:
        
        \( (1)(2) + (1)(2i) + (-2i)(2) + (-2i)(2i) = 2 + 2i - 4i - 4i^2 = 2 - 2i - 4(-1) = 2 - 2i + 4 = 6 - 2i \)
        
        And the right side is \( \frac{20}{3 + i} = 6 - 2i \), as we calculated earlier.
        
        \colorbox{Turquoise}{So, both sides match, confirming that \( z = 1 + 2i \) is indeed the correct solution.}
        
        \textbf{Final Answer}
        
        \[ \boxed{1 + 2i} \]
    \end{AIbox}
\end{figure*}

\begin{figure*}[!ht] 
    \begin{AIbox}{Qwen2.5-32B-Instruct-Preview Response Question 3}
        To solve the problem, let's first set up the equation and use the given conditions to find the complex number \(z\).

        Given: \(\overline{z}(z+1)=\frac{20}{3+i}\)
        
        Let's start by simplifying the right-hand side of the equation:
        
        \[
        \frac{20}{3+i} = \frac{20(3-i)}{(3+i)(3-i)} = \frac{20(3-i)}{3^2 - i^2} = \frac{20(3-i)}{9+1} = \frac{20(3-i)}{10} = 2(3-i) = 6 - 2i
        \]
        
        So, the equation becomes:
        
        \[
        \overline{z}(z+1) = 6 - 2i
        \]
        
        Let \(z = x + yi\) where \(x, y \in \mathbb{R}\) and \(x > 0\). Then, \(\overline{z} = x - yi\).
        
        Substituting \(z\) and \(\overline{z}\) into the equation:
        
        \[
        (x - yi)((x + yi) + 1) = 6 - 2i
        \]
        
        Simplifying the left-hand side:
        
        \[
        \begin{aligned}
            (x - yi)(x + 1 + yi) &= (x - yi)(x + 1 + yi) \\
            & = x(x + 1) + xyi - yix - yi^2 + (-yi)(yi) \\
            & = x(x + 1) + y^2 + xyi - xyi
        \end{aligned}
        \]
        
        Since \(i^2 = -1\), we get:
        
        \[
        x(x + 1) + y^2 + xyi - xyi = x^2 + x + y^2
        \]
        
        So, the equation becomes:
        
        \[
        x^2 + x + y^2 = 6 - 2i
        \]
        
        Separating the real and imaginary parts, we have:
        
        \[
        x^2 + x + y^2 = 6 \quad \text{and} \quad 0 = -2i
        \]
        
        Since the imaginary part of the left-hand side is zero and the imaginary part of the right-hand side is \(-2i\), it must be that the imaginary part of \(z\) is such that it cancels out. This means \(y = -2\).
        
        Substituting \(y = -2\) into the real part equation:
        
        \[
        x^2 + x + (-2)^2 = 6
        \]
        
        \[
        x^2 + x + 4 = 6
        \]
        
        \[
        x^2 + x - 2 = 0
        \]
        
        Solving the quadratic equation \(x^2 + x - 2 = 0\):
        
        \[
        x = \frac{-1 \pm \sqrt{1 + 8}}{2} = \frac{-1 \pm 3}{2}
        \]
        
        So, \(x = 1\) or \(x = -2\). Since the real part of \(z\) is greater than 0, we have \(x = 1\).
        
        Thus, \(z = 1 - 2i\).
        
        The solution is:
        \[
        \boxed{1 - 2i}
        \]

    \end{AIbox}
\end{figure*}

\clearpage

\begin{figure*}[t]
    \centering
    \includegraphics[scale=.5]{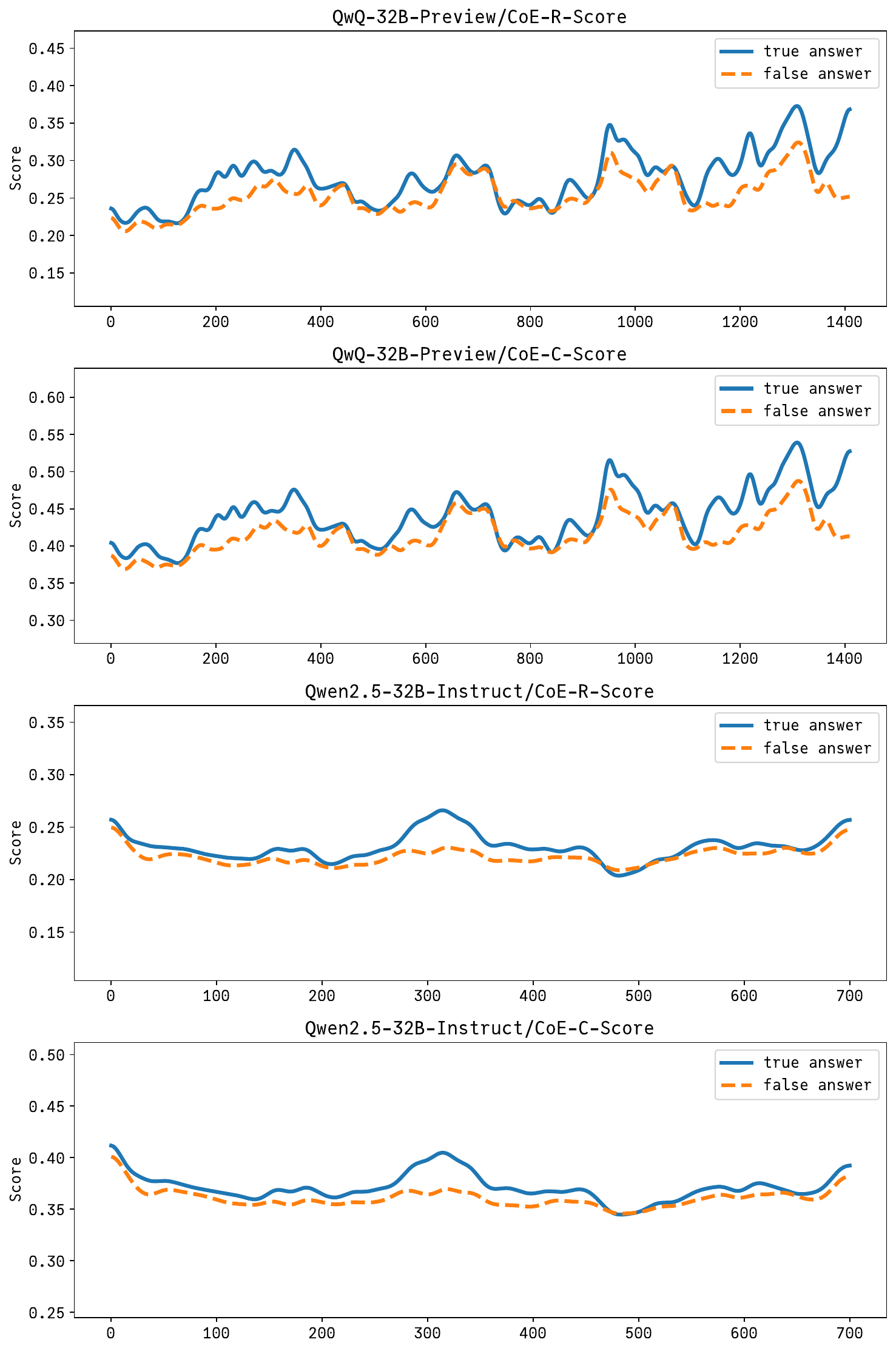}
    \caption{CoE-Score of QwQ and Qwen on Question 3.}
    \label{fig:coe_score3}
\end{figure*}

\begin{figure*}[!ht] 
    \begin{AIbox}{Question 4}
        Integers $a$ and $b$ are randomly chosen without replacement from the set of integers with absolute value not exceeding $10$. What is the probability that the polynomial $x^3 + ax^2 + bx + 6$ has $3$ distinct integer roots?
        \par
    \end{AIbox}
\end{figure*}

\begin{figure*}[!ht] 
    \begin{AIbox}{QwQ-32B-Preview Response for Question 4}

        So I've got this probability problem here. It says that integers \(a\) and \(b\) are randomly chosen without replacement from the set of integers with absolute value not exceeding 10. So, first things first, I need to understand what that set is. If the absolute value of the integers doesn't exceed 10, that means they can range from -10 to 10, inclusive. So the set is \(\{-10, -9, -8, \dots, 9, 10\}\). That's 21 integers in total.

        Now, we're choosing \(a\) and \(b\) without replacement from this set, meaning that \(a\) and \(b\) are distinct integers from this set. So the total number of possible pairs \((a, b)\) is \(21 \times 20 = 420\), since there are 21 choices for \(a\) and then 20 remaining choices for \(b\).
        
        The problem is asking for the probability that the polynomial \(x^3 + a x^2 + b x + 6\) has 3 distinct integer roots. So, for the polynomial to have 3 distinct integer roots, it must be factorable into \((x - r)(x - s)(x - t)\), where \(r, s, t\) are distinct integers.
        
        Let's expand \((x - r)(x - s)(x - t)\) to see how the coefficients relate to \(a\) and \(b\):
        
        \[
        (x - r)(x - s)(x - t) = x^3 - (r + s + t)x^2 + (rs + rt + st)x - rst
        \]
        
        Comparing this to the given polynomial \(x^3 + a x^2 + b x + 6\), we can see that:
        
        \[
        \begin{cases}
        -(r + s + t) = a \\
        rs + rt + st = b \\
        - r s t = 6
        \end{cases}
        \]
        
        ............................
        
        All these calculations seem correct.

        Now, is there any possibility that different sets of roots could lead to the same \((a, b)\) pair? For example, if two different triplets of roots result in the same \(a\) and \(b\), then that would mean that pair is counted only once in the favorable outcomes.
        
        \colorbox{Turquoise}{But in our case, each set of roots gives a distinct \((a, b)\) pair, as we can see:}
        
        - \((0, -7)\)
        
        - \((-2, -5)\)
        
        - \((-6, -1)\)
        
        - \((-4, 1)\)
        
        No overlaps here.
        
        Therefore, the number of favorable outcomes is indeed 4.
        
        Total possible outcomes are \(21 \times 20 = 420\).
        
        Thus, the probability is \(\frac{4}{420} = \frac{1}{105}\).
        
        \colorbox{Turquoise}{Wait, but perhaps I should consider that \(a\) and \(b\) are chosen without replacement}, meaning that \((a, b)\) and \((b, a)\) are different unless \(a = b\), which isn't the case here. But in our favorable outcomes, each \((a, b)\) is unique and distinct from \((b, a)\), since none of the \(a\) and \(b\) pairs have \(a = b\).
        
        Therefore, no need to adjust for that.
        
        \colorbox{Turquoise}{Alternatively, if the problem considered \((a, b)\) and \((b, a)\) as the same,} but since it's about choosing without replacement and the polynomial has distinct roots, I think the ordered pairs are appropriate here.
        
        So, I think \(\frac{1}{105}\) is the correct probability.
        
        \textbf{Final Answer}
        
        \[ \boxed{\dfrac{1}{105}} \]
    \end{AIbox}
\end{figure*}

\begin{figure*}[!ht] 
    \begin{AIbox}{Qwen2.5-32B-Instruct-Preview Response Question 4}
        To solve this problem, we need to analyze the conditions under which the polynomial \(x^3 + ax^2 + bx + 6\) has three distinct integer roots. 

        1. \textbf{Identify the Set of Integers:}
           The set of integers with absolute value not exceeding \(10\) is \(\{-10, -9, -8, \ldots, 9, 10\}\). There are \(21\) integers in this set.
        
        2. \textbf{Polynomial Roots:}
           Let the roots of the polynomial \(x^3 + ax^2 + bx + 6\) be \(r, s,\) and \(t\). By Vieta's formulas, we know:
           \[
           r + s + t = -a, \quad rs + rt + st = b, \quad rst = -6
           \]
        
        3. \textbf{inding Integer Roots:}
           The integer roots \(r, s,\) and \(t\) must satisfy \(rst = -6\). We list the possible sets of integer roots:
           \[
           \{1, 2, -3\}, \{1, -2, 3\}, \{-1, 2, 3\}, \{-1, -2, -3\}
           \]
        
        4. \textbf{Counting Valid Combinations:}
           We need to count the number of ways to choose \(a\) and \(b\) such that the polynomial has these roots. Each set of roots \(\{r, s, t\}\) must be chosen from the set \(\{-10, -9, \ldots, 9, 10\}\) without replacement.
        
        5. \textbf{Calculate \(a\) and \(b\) for Each Set:}
           For each set of roots, calculate \(a\) and \(b\):
           - For \(\{1, 2, -3\}\):
             \[
             r + s + t = 1 + 2 - 3 = 0 \implies a = 0
             \]
             \[
             rs + rt + st = 1 \cdot 2 + 1 \cdot (-3) + 2 \cdot (-3) = 2 - 3 - 6 = -7 \implies b = -7
             \]
        
           - For \(\{1, -2, 3\}\):
             \[
             r + s + t = 1 - 2 + 3 = 2 \implies a = -2
             \]
             \[
             rs + rt + st = 1 \cdot (-2) + 1 \cdot 3 + (-2) \cdot 3 = -2 + 3 - 6 = -5 \implies b = -5
             \]
        
           - For \(\{-1, 2, 3\}\):
             \[
             r + s + t = -1 + 2 + 3 = 4 \implies a = -4
             \]
             \[
             rs + rt + st = (-1) \cdot 2 + (-1) \cdot 3 + 2 \cdot 3 = -2 - 3 + 6 = 1 \implies b = 1
             \]
        
           - For \(\{-1, -2, -3\}\):
             \[
             r + s + t = -1 - 2 - 3 = -6 \implies a = 6
             \]
             \[
             rs + rt + st = (-1) \cdot (-2) + (-1) \cdot (-3) + (-2) \cdot (-3) = 2 + 3 + 6 = 11 \implies b = 11
             \]
        
        6. \textbf{Total Possible Choices for \(a\) and \(b\):}
           The total number of ways to choose \(a\) and \(b\) from the set of integers \(\{-10, -9, \ldots, 9, 10\}\) without replacement is:
           \[
           \binom{21}{2} = \frac{21 \times 20}{2} = 210
           \]
        
        7. \textbf{Probability Calculation:}
           There are \(4\) valid combinations of \((a, b)\) that make the polynomial have three distinct integer roots. Therefore, the probability is:
           \[
           \frac{4}{210} = \frac{2}{105}
           \]
        
        Thus, the probability that the polynomial \(x^3 + ax^2 + bx + 6\) has three distinct integer roots is \(\boxed{\frac{2}{105}}\).

    \end{AIbox}
\end{figure*}

\clearpage

\begin{figure*}[t]
    \centering
    \includegraphics[scale=.5]{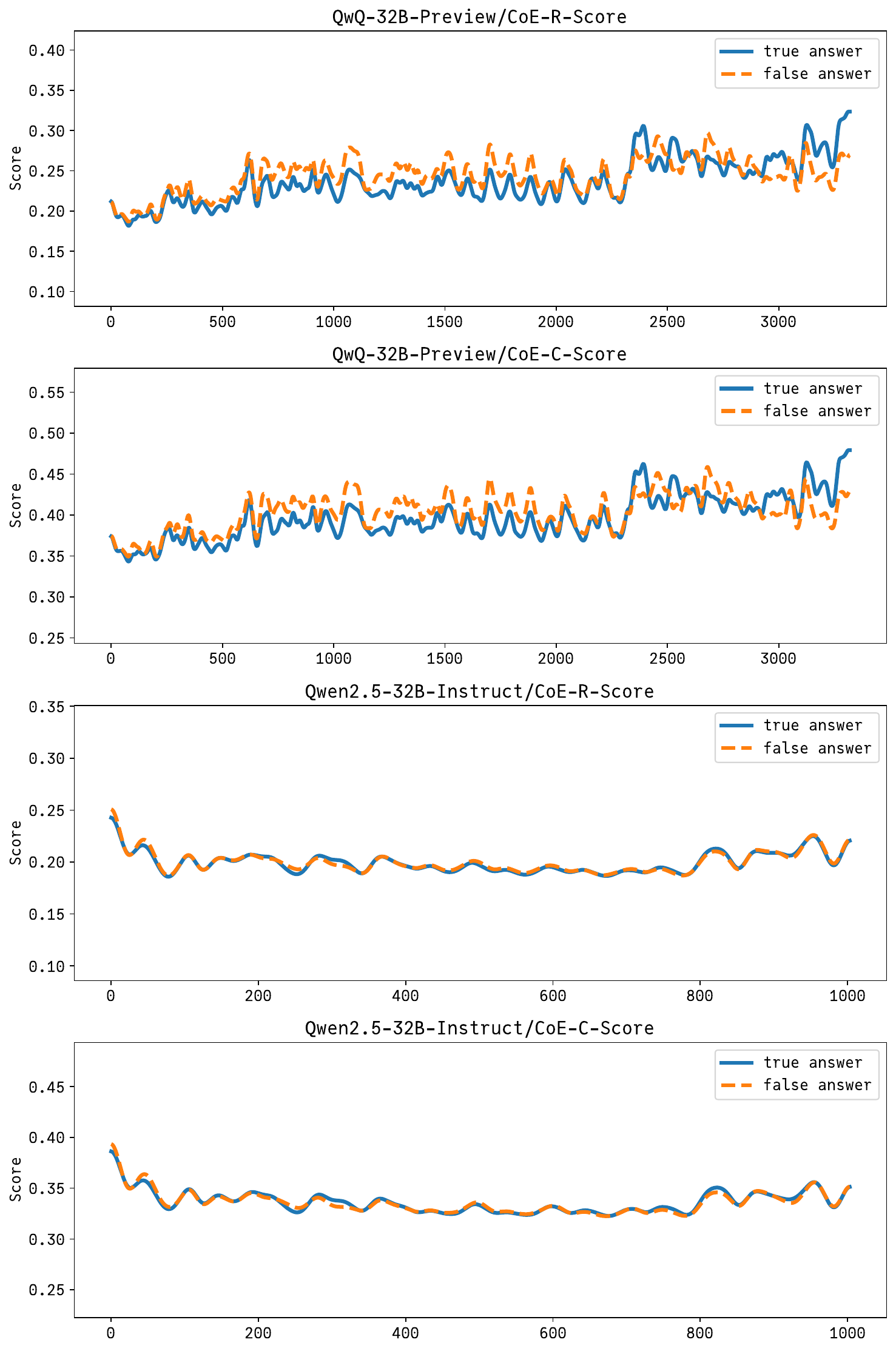}
    \caption{CoE-Score of QwQ and Qwen on Question 4.}
    \label{fig:coe_score4}
\end{figure*}

\end{document}